\documentclass{article}

     \PassOptionsToPackage{numbers, compress}{natbib}

     \usepackage[final]{neurips_2022}

\usepackage[utf8]{inputenc} 
\usepackage[T1]{fontenc}    
\usepackage{hyperref}       
\usepackage{url}            
\usepackage{booktabs}       
\usepackage{amsfonts}       
\usepackage{nicefrac}       
\usepackage{microtype}      
\usepackage{xcolor}         
\usepackage{enumitem}

\usepackage{amsthm}
\usepackage{verbatim}
\usepackage{bbm} 
\usepackage{mathtools}

\usepackage{wrapfig}
\usepackage{multirow}

\usepackage{multibib}
\newcites{Supp}{Supplement References}

\newcommand{\review}[1]{\textcolor{black}{#1}}

\newcommand{\M}{\mathcal{M}}
\newcommand{\Z}{\mathcal{Z}}
\newcommand{\mP}{\mathbf{P}}
\newcommand{\mQ}{\mathbf{Q}}
\newcommand{\mK}{\mathbf{K}}

\newcommand{\mM}{\mathbf{M}}
\newcommand{\mD}{\mathbf{D}}
\newcommand{\mpi}{\boldsymbol{\pi}}
\newcommand{\X}{\mathsf{X}}
\newcommand{\hX}{\widehat{\mathsf{X}}}
\newcommand{\cX}{\mathcal{X}}
\newcommand{\real}{\mathbb{R}}

\newcommand{\mmu}{\boldsymbol{\mu}}

\usepackage{amsthm,amsmath,amsfonts,amssymb}
\usepackage{bbm}
\theoremstyle{plain}

\newtheorem{thm}{Theorem}
\theoremstyle{definition}
\newtheorem{definition}{Definition} 

\newtheorem{cor}{Corollary}

\newtheorem{innercustomthm}{Theorem}
\newenvironment{customthm}[1]
  {\renewcommand\theinnercustomthm{#1}\innercustomthm}
  {\endinnercustomthm}
\usepackage{algorithm}
\usepackage{algpseudocode}

\title{Manifold Interpolating Optimal-Transport Flows for Trajectory Inference}

\author{
Guillaume Huguet$^1$\thanks{Equal contribution} \quad D.S. Magruder$^2$\footnotemark[1] \quad Alexander Tong$^1$\footnotemark[1] \quad Oluwadamilola Fasina$^2$  \quad
\AND
Manik Kuchroo$^2$ \quad Guy Wolf$^{1}$\thanks{Corresponding authors: \texttt{guy.wolf@umontreal.ca} and \texttt{smita.krishnawamy@yale.edu}} \quad Smita Krishnaswamy$^{2}$\footnotemark[2]\vspace{4pt}\\
$^1$Université de Montréal; Mila - Quebec AI Institute \quad $^2$ Yale University}

\begin{document}

\maketitle

\begin{abstract}
We present a method called Manifold Interpolating Optimal-Transport Flow (MIOFlow) that learns stochastic, continuous population dynamics from static snapshot samples taken at sporadic timepoints. MIOFlow combines dynamic models,  manifold learning, and optimal transport by training neural ordinary differential equations (Neural ODE) to interpolate between static population snapshots as penalized by optimal transport with manifold ground distance. Further, we ensure that the flow follows the geometry by operating in the latent space of an autoencoder that we call a geodesic autoencoder (GAE). In GAE the latent space distance between points is regularized to match a novel multiscale geodesic distance on the data manifold that we define. We show that this method is superior to normalizing flows, Schr\"odinger bridges and other generative models that are designed to flow from noise to data in terms of interpolating between populations. Theoretically, we link these trajectories with dynamic optimal transport. We evaluate our method on simulated data with bifurcations and merges, as well as scRNA-seq data from embryoid body differentiation, and acute myeloid leukemia treatment. 
\end{abstract}

\section{Introduction}
Here, we tackle the problem of continuous dynamics of probability distributions defined on a data manifold. Data from naturalistic systems are often modeled as generated from an underlying low dimensional manifold embedded in a high dimensional measurement space. Termed the manifold hypothesis, this assumption has led to many successful models of biological, chemical, and physical systems. Measurements in such systems are increasingly high dimensional. For instance, in single cell data the entire transcriptomic profile of the cell is measured with each mRNA or gene species a dimension. However, because of informational redundancy between these genes, the intrinsic dimensionality of the data is low dimensional. Mathematically, a Riemannian manifold is a good model for such a system, and much of manifold learning literature including, diffusion maps~\cite{coifman_diffusion_2006}, Laplacian eigenmaps~\cite{belkin_semi-supervised_2004}, and the wider graph signal processing literature, has focused on learning this structure via data graphs~\cite{moon_manifold_2018, moon_visualizing_2019} and autoencoders~\cite{mishne2019diffusion,duque2020extendable}.

Building on this literature, we consider snapshot measurements of cellular populations over time. Current technology cannot follow a cell in time as the measurements are destructive. We frame this as learning a population flow on a data manifold. Recently several neural networks that perform flows, or transports between probability distributions, have been proposed in the literature. However, most of these works have focused on generative modeling: i.e., flowing from a noise distribution such as a Gaussian distribution to a data distribution in order to generate data. Examples include score-based generative matching~\cite{song_generative_2019, song2020score}, diffusion models~\cite{ho_denoising_2020}, Schr\"odinger bridges~\cite{pavon2021data,de2021diffusion}, and continuous normalizing flows (CNF)~\cite{chen2018neural,grathwohl_ffjord:_2019}. However, here we focus on learning continuous dynamics of such systems using static snapshot measurements, with a key task being interpolation in unmeasured times, as well as inference of individual trajectories that follow the manifold structure.

To continuously interpolate populations over time on data with manifold structure, we propose Manifold Interpolation Optimal-transport Flow (MIOFlow)\footnote{Code is available here: \url{https://github.com/KrishnaswamyLab/MIOFlow}}, a new framework for flows based on dynamical optimal transport in a manifold embedding. MIOFlow uses a neural ordinary differential equations (Neural ODE)~\cite{chen2018neural} to transport a sampling of high dimensional data points between timepoints such that 1.\ the transport occurs on a manifold defined by samples, 2.\ the transport is penalized to agree with measured timepoints using Wasserstein, 3.\ the transport is inherently stochastic. 

Works tackling a similar problem include TrajectoryNet~\cite{tong_trajectorynet_2020}, dynamics modeled by a recurrent neural network~\cite{hashimoto2016learning}, and Waddington-OT~\cite{schiebinger_reconstruction_2017}. TrajectoryNet is based on a continuous normalizing flow where the magnitude of the derivative is penalized to create a more optimal flow. However, this approach suffers from several drawbacks. The first drawback is the requirement of starting from a Gaussian distribution. It can be difficult to match distributions from the real world to Gaussian distributions in an interpolating sense, all intermediate distributions must be penalized to start from a Gaussian rather than being trained to flow from one to the next. Second, continuous normalizing flows are deterministic. Hence, to model the intrinsic stochasticity in biology, we have to force chaotic dynamics starting from slight noise added to any initial distributions. Third, continuous normalizing flow models in $k$ dimensions require calculating the trace of the Jacobian, which requires $O(k^2)$ operations to compute~\cite{chen2018neural}, making our method $k$ times faster per function evaluation.

Additionally, unlike normalizing flows which operate in ambient data dimensions, we focus on flows on a data manifold. We feature a two-pronged approach with which to enforce this.  First, propose the {\em Geodesic Autoencoder} to embed the data such that distances in the latent space match a new manifold distance called diffusion geodesic distance. Second, we penalize the transport using a manifold optimal transport method~\cite{peyre_computational_2020,villani2009optimal}.  These two steps enforce flows on the manifold, whereas in earlier works, such as TrajectoryNet, the KL-divergence is used to match distributions, and in \cite{mathieu2020riemannian} access to the metric tensor of a Riemannian manifold is required.

The main contributions of our work include: 
\begin{itemize}[itemsep=0pt,parsep=0pt,leftmargin=15pt]
    \item The MIOFlow framework for efficiently learning continuous stochastic dynamics of static snapshot data based on Neural ODEs that implements optimal transport flows on a data manifold.
    \item The Geodesic Autoencoder to create a manifold embedding in the latent space of the autoencoder. 
    \item A new multiscale manifold distance called diffusion geodesic distance, and theoretical results showing its convergence to a geodesic on the manifold in the limit of infinitely many points. 
    \item Empirical results showing that our flows can model divergent trajectories on toy data and on single-cell developmental and treatment response data. 
\end{itemize}

\vspace{-4mm}
\section{Preliminaries and Background}\label{sec: preliminaries}
\paragraph{Problem Formulation and Notation}
We consider a distribution $\mu_t$ over $\real^k$ evolving over time $t\in\real$, from which we only observe samples from a finite set of $T$ distributions $\{\mu_i\}_{i=0}^{T-1}$. For each time $t$, we observe a sample $\X_t\sim\mu_t$ of size $n_t$. We note the set of all observations $\X$ of size $n:=\sum_i n_i$. We also want to characterize the evolution of the support of $\mu_t$. We aim to define a trajectory from an initial points $X_0\sim\mu_0$ to $X_{T-1}\sim\mu_{T-1}$, given the intermediate conditions $X_2\sim\mu_2, \dotsc, X_{T-1}\sim\mu_{T-1}$. We are thus interested in matching distributions given a set of $T$ samples $\{\X_i\}_{i=1}^{T-1}$, and initial condition $\X_0$. 

In the following, we assume that the distributions are absolutely continuous with respect to the Lebesgue measure, and we use the same notation for the distribution and its density. We note the equivalence between two distances $d_1\simeq d_2$. We assume that all Stochastic Differential Equations (SDE) admit a solution~\cite{gardiner1985handbook}. All proofs are presented in the supplementary material.

\paragraph{Optimal Transport}
In this section, we provide a brief overview of optimal transport~\cite{peyre_computational_2020,villani2009optimal}, our primary approach for interpolating between distributions using a neural ODE. We consider two distributions $\mu$ and $\nu$ defined on $\cX$ and $\mathcal{Y}$, and $\Pi(\mu,\nu)$ the set of joint distributions on $\cX\times\mathcal{Y}$ with marginals $\mu$ and $\nu$, i.e. $\pi(dx,\mathcal{Y}) = \mu(dx)$ and $\pi(\cX,dy) = \nu(dy)$ for $\pi\in\Pi(\mu,\nu)$. The transport plan $\pi$ moves the mass from $\mu$ to $\nu$, where the cost of moving a unit mass from the initial $x\in\cX$ to the final $y\in\mathcal{Y}$ is $d(x,y)$. This formulation gives rise to the $p$-Wasserstein distance
\begin{equation*}
    W_p(\mu, \nu)^p := \inf_{\pi \in \Pi(\mu, \nu)} \int_{\cX \times \mathcal{Y}} d(x, y)^p \pi(dx, dy),
\end{equation*}
where $p\in[1,\infty)$. Suppose $\cX = \mathcal{Y} = \real^k$, and $d(x,y) := ||x-y||_2$ then \citet{benamou_computational_2000} provide a \textit{dynamic formulation} of the optimal transport problem. For simplicity, we assume a fix time interval $[0,1]$, but note that the following holds for any time interval $[t_0,t_1]$. The transport plan is replaced by a time-evolving distribution $\rho_t$, such that $\rho_0=\mu$ and $\rho_1=\nu$ and that satisfies the continuity equation $\partial_t \rho_t + \nabla \cdot (\rho_t v) = 0$, where $\nabla \cdot$ is the divergence operator. The movement of mass is described by a time-evolving vector field $v(x,t)$. When $\rho_t$ satisfies these conditions, \citet{benamou_computational_2000} show that 
\begin{equation}\label{eq: dyn_w2}
    W_2(\mu, \nu)^2 = \inf_{(\rho_t,v)}\int_{0}^{1}\int_{\real^k} \|v(x,t)\|_2^2 \rho_t(dx) dt.
\end{equation}
This formulation arises from the field of fluid mechanics; the optimal vector field is the divergence of a pressureless potential flow. We can also view the problem on the path space of $\real^k$
\begin{equation}\label{eq: path_w2}
    W_2(\mu,\nu)^2 = \inf_{X_t}  \mathbb{E}\bigg[ \int_0^1 \|f(X_t,t)\|_2^2 dt \bigg] \, \text{ s.t. }dX_t = f(X_t,t)dt,\, X_0\sim\mu,\,X_1\sim\nu,
\end{equation}
where the infimum is over all absolutely continuous stochastic path $X_t$ (see \cite{mikami2004monge}).

\paragraph{Adding Diffusion}

For various applications, it is interesting to incorporate a diffusion term in the paths or trajectories. These types of flows are often used in control dynamics~\cite{agrachev2010continuity,mikami2006duality,ghoussoub2021solution}, and mean field games~\cite{lasry2007mean}. We consider an SDE $dX_t = f(X_t,t) dt + \sqrt{\sigma}dB_t$, where $B_t$ is a standard Brownian motion. To model this new dynamic, one can replace the continuity equation with the Fokker–Planck equation of this SDE with diffusion, i.e. $\partial_t\rho_t + \nabla \cdot (\rho_t v) = \Delta(\sigma\rho_t/2)$. From this formulation, one can also retrieve the Benamou-Brenier optimal transport. Indeed, \citet{mikami2004monge} established the convergence to the $W_2$ when the diffusion term goes to zero. Similar to \eqref{eq: path_w2}, we can phrase the problem using an SDE
\begin{equation}\label{eq: inf SDE_diffusion}
    \inf_{f} \mathbb{E}\bigg[ \int_0^1 \|f \|_2^2dt \bigg] \text{ s.t. } dX_t = f(X_t,t) dt + \sqrt{\sigma}dB_t, \, X_0\sim\mu, \, X_1\sim\nu.
\end{equation}
The two previous formulations admit the same entropic interpolation $\rho_t$~\cite{pavon2021data,mikami2004monge}.
In this paper, we utilize such a diffusion term, with the knowledge that it still converges to the  transport between distributions. Moreover, as the diffusion goes to zero, the infimum converges to the $W_2$. 


\paragraph{Manifold Learning}\label{sec: manifold_learning}
A useful assumption in representation learning is that high dimensional data originates from an intrinsic low dimensional manifold that is mapped via nonlinear functions to observable high dimensional measurements; this is commonly referred to as the manifold assumption. Formally, let $\mathcal{M}$ be a hidden $m$ dimensional manifold that is only observable via a collection of $k \gg m$ nonlinear functions $f_1,\ldots,f_k : \mathcal{M} \to \real$ that enable its immersion in a high dimensional ambient space as $F(\mathcal{M}) = \{\mathbf{f}(z) = (f_1(z),\ldots,f_k(z))^T : z \in \mathcal{M} \} \subseteq \real^k$ from which data is collected. Conversely, given a dataset $\X = \{x_1, \ldots, x_n\} \subset \real^k$ of high dimensional observations, manifold learning methods assume data points originate from a sampling $Z = \{z_i\}_{i=1}^n \in \mathcal{M}$ of the underlying manifold via $x_i = \mathbf{f}(z_i)$, $i = 1, \ldots, n$, and aim to learn a low dimensional intrinsic representation that approximates the manifold geometry of~$\mathcal{M}$. 

To learn a manifold geometry from collected data
that is robust to sampling density variations, \citet{coifman_diffusion_2006} proposed to use an anisotropic kernel $k_{\epsilon,\beta}(x,y) := k_\epsilon(x,y)/\|k_\epsilon(x,\cdot)\|_1^\beta \|k_\epsilon(y,\cdot)\|_1^{\beta}$, where $0 \leq \beta \leq 1$ controls the separation of geometry from density, with $\beta = 0$ yielding the isotropic kernel, and $\beta = 1$ completely removing density and providing a geometric equivalent to uniform sampling of the underlying manifold. Next, the kernel $k_{\epsilon,\beta}$ is normalized to define transition probabilities $p_{\epsilon,\beta}(x,y) := k_{\epsilon,\beta}(x,y)/\|k_{\epsilon,\beta}(x,\cdot)\|_1$ and an $n \times n$ row stochastic matrix $(\mathbf{P}_{\epsilon,\beta})_{ij} := p_{\epsilon,\beta}(x_i, x_j)$ 
that describes a Markovian diffusion process over the intrinsic geometry of the data. Finally, a diffusion map~\cite{coifman_diffusion_2006} $\Phi_t(x_i)$ is defined by the eigenvalues and eigenvectors of the matrix $\mP^t_{\epsilon,\beta}$. Most notably, this embedding preserves the diffusion distance $\|p^t_{\epsilon,\beta}(x_i,\cdot)-p^t_{\epsilon,\beta}(x_j,\cdot)/\phi_1(\cdot)\|_2$ between $x_i,x_j\in\X$, where $t$ is a time-scale parameter. Next, we consider a multiscale diffusion distance that relates to the geodesic distance on the manifold.


\begin{figure}[t]
    \centering
    \includegraphics[width=0.7\linewidth]{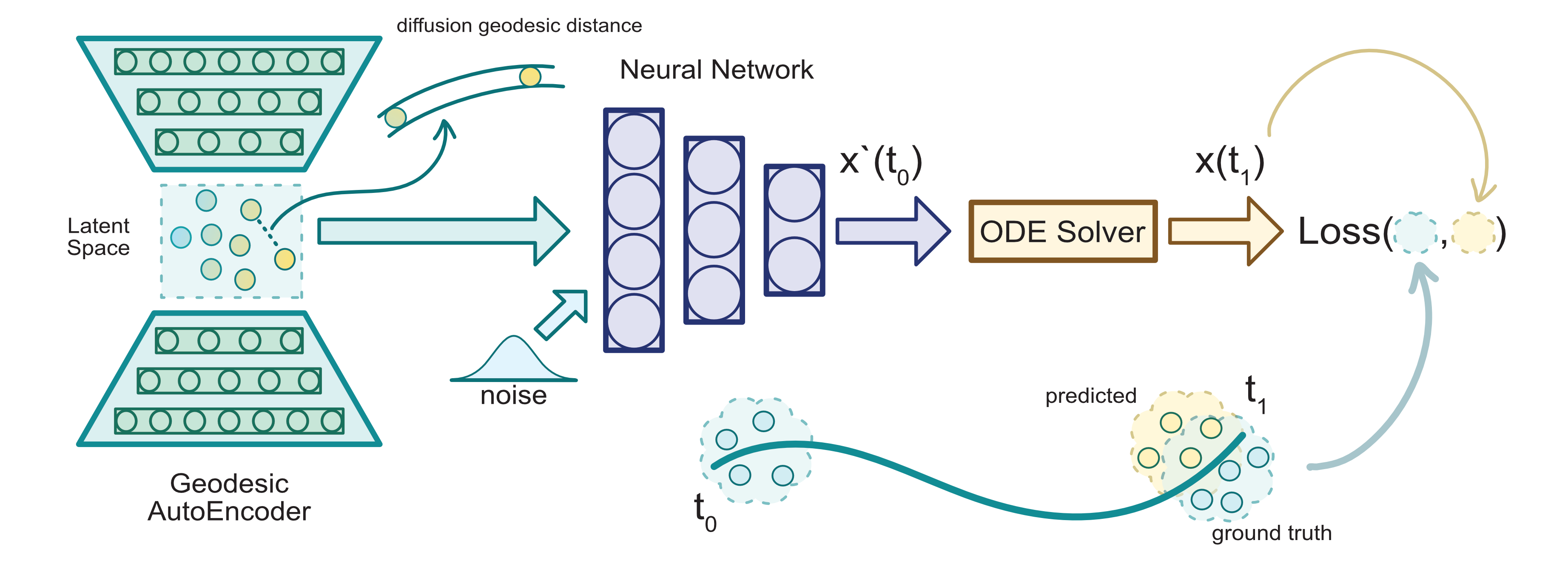}
    \vspace{-2mm}
    \caption{Overview of the MIOFlow pipeline. The Geodesic Autoencoder learns a latent space that preserves the diffusion geodesic distance. A neural network predicts the derivative of the trajectories with respect to time. For an initial sample at $t_0$, the ODE Solver produces the trajectories $x(t_1)$. At a population level, predictions of points at time $t_1$ are penalized by the Wasserstein distance during training.}
    \label{fig:schematic}
    \vspace{-4mm}
\end{figure}

\paragraph{Manifold Geodesics}\label{sec: manifold geodesics}
The dynamic optimal transport formulations \eqref{eq: dyn_w2} and \eqref{eq: path_w2} are valid when the ground distance is Euclidean. This is restrictive as we might have observations sampled from a lower dimensional manifold. Thus, using the Euclidean distance in the ambient space ignores the underlying geometry of the data. Static optimal transport with a geodesic distance is known to perform well on geometric domains~\cite{solomon2015convolutional,tong2022embedding}, or when the data lie on a manifold~\cite{tong2021diffusion}. Here, we extend the use of a ground geodesic distance to the dynamic formulation. To do so, we define a manifold learning step that precedes the trajectory inference. The goal is to learn an embedding $\Z$ such that the Euclidean in $\Z$ is equivalent to the geodesic distance. In that case, dynamic optimal transport in $\Z$ is equivalent to the Wasserstein with a geodesic ground distance. In the following section, we present theoretical results that will justify our approximation of a geodesic distance on a closed Riemannian manifold.

We consider a closed Riemannian manifold  $(\M,d_\M)$, where $d_\M$ is the geodesic distance \review{representing the shortest path between two points on the manifold.} We note $h_t$ the heat kernel on $\M$. For a point $x\in\M$, the heat kernel $h_t(x,\cdot)$ induces a measure, i.e. how the heat has propagated on the manifold at time $t$ given an arbitrary initial distribution. The diffusion ground distance between $x,y\in\M$ is based on the $L^1$ norm between the measures induced by the heat kernel given the initial conditions $\delta_x$ and $\delta_y$.
\begin{definition}
\label{def:diffusion_alpha}
The diffusion ground distance between $x,y \in \mathcal{M}$ is
\begin{equation*}
    D_\alpha(x,y) := \sum_{k\geq 0}2^{-k
    \alpha}||h_{2^{-k}}(x,\cdot)-h_{2^{-k}}(y,\cdot)||_1,
\end{equation*}
for $\alpha\in(0,1/2)$, the scale parameter $k\geq0$, and $h_t$ the heat kernel on $\M$.
\end{definition}

Next, we state an important result from Leeb and Coifman~(Thm.~2 in~\cite{leeb_holderlipschitz_2016}). This theorem links the diffusion ground distance and the geodesic on a Riemannian manifold. 
\begin{thm}[\cite{leeb_holderlipschitz_2016} Thm.~2]\label{thm: geodesic_Leeb} Let $(\M,d_\M)$ a closed Riemannian manifold, with geodesic $d_\M$, for $\alpha\in(0,1/2)$, the distance $D_\alpha$ is equivalent to $d_\M^{2\alpha}$.
\end{thm}
In practice, we cannot always evaluate the heat kernel for an arbitrary manifold. However, convergence results from~\citet{coifman_diffusion_2006} provide a natural approximation. \review{Indeed, for $\beta=1$, the diffusion operator $P^{t/\epsilon}_{\epsilon,\beta}$ converges to the heat operator as $\epsilon$ goes to zero (Prop.\ 3~\cite{coifman_diffusion_2006}). For the rest of the paper we assume $\beta =1$, we define $P_\epsilon := P_{\epsilon,1}$ and the matrix $\mP_\epsilon:=\mP_{\epsilon,1}$. } 

\section{Manifold Interpolating Optimal-Transport Flow}\label{sec:MIOFlow}

To learn individual trajectories from multiple cross-sectional samples of a population, we propose MIOFlow shown in Fig. \ref{fig:schematic}. Our method consists of two steps. We first learn an embedding $\Z$ which preserves the diffusion geodesic distance---which we define using our Geodesic Autoencoder. Then, we learn continuous trajectories based on an optimal transport loss. We model the trajectories with a Neural ODE~\cite{chen2018neural}, allowing us to learn non-linear paths and interpolate between timepoints.

\subsection{Geodesic Autoencoder Embedding}\label{sec: manifold_reg} 

In light of Thm.~\ref{thm: geodesic_Leeb}, we define an approximation of the diffusion ground distance $D_\alpha$ (Def.~\ref{def:diffusion_alpha}) from a collection of samples $\X$ that we call {\em diffusion geodesic distance}. \review{We then train an autoencoder regularized to match this distance. Similar ideas are explored in \cite{mishne2019diffusion,duque2020extendable} where the latent space of the autoencoder is regularized to match a manifold learning embedding, either PHATE or diffusion map.}

Our approximation relies on \review{ the diffusion matrix $\mP_\epsilon$. We first recall its construction.} We build a graph with the affinity matrix defined by $(\mK_\epsilon)_{ij}:= k_\epsilon(x_i,x_j)$, and we consider the density normalized matrix $\mM_\epsilon:= \mQ^{-1} \mK_\epsilon\mQ^{-1}$, where $\mQ$ is a diagonal matrix with $\mQ_{ii} := \sum_j (\mK_\epsilon
)_{ij}$. In practice, we choose $k_\epsilon$ to be a Gaussian kernel or the $\alpha$-decay kernel \cite{moon_visualizing_2019}. Lastly, a Markov diffusion operator is defined by $\mP_\epsilon := \mD^{-1}\mM_\epsilon$, where $\mD_{ii} := \sum_{j=1}^n (\mM_\epsilon)_{ij}$ is a diagonal matrix. The stationary distribution associated to $\mP_\epsilon$ is $\mpi$, where $\mpi_i = \mD_{ii}/\sum_j \mD_{jj}$, since $\mP_\epsilon$ is $\mpi$-reversible. Finally, we note that the matrix $\mP_\epsilon$ follows a similar construction as the kernel $p_\epsilon$; the integrals are approximated by sums. We refer to~\cite{coifman_diffusion_2006} for more details about the convergence of the matrix operator $\mP_\epsilon$ to the operator $P_\epsilon$. Until now, we have defined a matrix $\mP_\epsilon$ that approximates the operator $P_\epsilon$, which in turn converges to the heat operator. Now, we define an approximation of the diffusion ground distance, based on the matrix $\mP_\epsilon$. We use the notation $(\mP_\epsilon)^t_{i:}$ to represent the i-th row of $\mP^t_\epsilon$, it represents the transition probabilities of a t-steps random walk started at $x_i$. 

\begin{definition}\label{def:approx_diff_ground}
We define the {\em diffusion geodesic distance} between $x_i, x_j \in \X$ as
\begin{equation*}
    G_\alpha(x_i,x_j) := \sum_{k=0}^K 2^{-(K-k)\alpha} || (\mP_\epsilon)_{i:}^{2^k} - (\mP_\epsilon)_{j:}^{2^k} ||_1 + 2^{-(K+1)/2} ||\mpi_i - \mpi_j ||_1.
\end{equation*}
\end{definition}
The diffusion geodesic compares the transitions probabilities of a random walk at various scales given two different initial states $x_i,x_j$. \review{In \cite{tong2021diffusion}, the authors use the diffusion operator $\mP_\epsilon$ to define a distance equivalent to the Wasserstein with ground distance $D_\alpha$. Their method comes from the static formulation of optimal transport. Here we propose to learn a space $\Z$ that preserves an approximation of the distance $D_\alpha$, in order to do dynamic optimal transport in $\Z$.}

\paragraph{Training}To use the diffusion geodesic distance, we train a \emph{Geodesic Autoencoder}, with encoder outputting $\phi:\real^k\to\Z$, such that $\|\phi(x_i)-\phi(x_j)\|_2^2 \approx G_\alpha(x_i,x_j)$. We draw a subsample of size $N$, evaluate $G_\alpha$ and minimize the Mean Square Error (MSE)
\begin{equation*}
    L(\phi) := \frac{2}{N}\sum_{i=1}^N\sum_{j>i}\left( ||\phi(x_i)-\phi(x_j)||_2 - G_\alpha(x_i,x_j) \right)^2.
\end{equation*}

Learning the embedding $\Z$ over using $G_\alpha$ as the advantage of being inductive, it is useful since we use it to compute distances between predicted and ground truth observations. Moreover, the encoder can be trained to denoise, hence becoming robust to predicted values that are not close to the manifold. Computing $G_\alpha$ on the entire dataset is inefficient due to the powering of the diffusion matrix. We circumvent this difficulty with the encoder, since we train on subsamples. 
We choose $N$ to have few observations in most regions of the manifold, thus making the computation of $G_\alpha$ very efficient, which allows us to consider more scales.
\begin{thm}
Assuming $\X$ is sampled from a closed Riemannian manifold $\M$ with geodesic $d_\M$. Then, for all $\alpha\in(0,1/2)$, sufficiently large $K,N$ and small $\epsilon>0$, we have with high probability $G_\alpha(x_i,x_j) \simeq d_\M^{2\alpha}(x_i,x_j)$ for all $x_i,x_j\in\X$.
\end{thm}
\begin{cor}\label{cor:zeroloss_geo}
If the encoder is such that $L(\phi)=0$, then with high probability $\|\phi(x_i)-\phi(x_j)\|_2 \simeq d_\M^{2\alpha}(x_i,x_j)$ for all $x_i,x_j\in\X$.
\end{cor}
The stochasticity in the two previous results arises from the discrete approximation of the operator $P_\epsilon$. The law or large numbers guarantees the convergence. For fix sample size, there exist approximation error bounds with high probability, see for example \cite{cheng2022eigen,singer_graph_2006}. In practice, we choose $\alpha$ close to $1/2$ so that the diffusion distance is equivalent to the geodesic on the manifold. From that embedding $\Z$, we can also train a \emph{decoder} $\phi^{-1}: \Z\to\real^k$, with a reconstruction loss $L_r:= \sum_x \|\phi^{-1}\circ\phi(x)-x\|_2$. This is particularly interesting in high dimensions, since we can learn the trajectories in the lower dimensional space $\Z$, then decode them in the ambient space. 
For example, we use the decoder in Sec.~\ref{sec: results} to infer cellular trajectories in the gene space, enabling us to understand how specific genes evolved over time. We describe the training procedure of the GAE in algorithm~\ref{alg: gae} in the supplementary material.

\subsection{Inferring Trajectories} 

Given $T$ distributions $\{\mu_i\}_{i=0}^{T-1}$, we want to model the trajectories respecting the conditions $X_i\sim\mu_i$ for fix timepoints $i\in\{0,\dotsc,T-1\}$. Formally, we want to learn a parametrized function $f_\theta(x,t)$ such that 
\begin{equation}\label{eq: traj}
    X_t = X_0 + \int_0^t f_\theta(X_u,u)du,\;\text{with }X_0\sim\mu_0,\dotsc, X_{T-1}\sim\mu_{T-1}.
\end{equation}
We adapt the main theorem from \cite{tong_trajectorynet_2020}, to consider the path space, and other types of dissimilarities between distributions.
\begin{thm}
We consider a time-varying vector field $f(x,t)$ defining the trajectories $dX_t = f(X_t,t)dt$ with density $\rho_t$, and a dissimilarity between distributions such that $D(\mu,\nu)=0$ iff $\mu=\nu$. Given these assumptions, there exist a sufficiently large $\lambda>0$ such that
\begin{equation}\label{eq: W2_in_thm}
    W_2(\mu,\nu)^2 =\inf_{X_t}  E\bigg[ \int_0^1 \|f(X_t,t)\|_2^2 dt \bigg] + \lambda D(\rho_1,\nu) \, \text{ s.t. }\, X_0\sim\mu.
\end{equation}
Moreover, if $X_t$ is defined on the embedded space $\Z$, then $W_2$ is equivalent to the Wasserstein with geodesic distance $W_2(\mu,\nu)\simeq W_{d_\mathcal{M}^{2\alpha}}(\mu,\nu)$.
\end{thm}
This theorem enables us to add the second marginal constraint in the optimization problem. In practice, it justifies the method of matching the marginals and adding a regularization on the vector field $f(x,t)$, that way the optimal $(\rho_t,f)$ corresponds to the one from the $W_2$. If in addition we model the trajectories in the embed space $\Z$, then the transport is equivalent to the one on the manifold. 

\paragraph{Training}We observe discrete distributions $\mmu_i := (1/n_i)\sum_i\delta_{x_i}$ for $x_i\in\X_i$, and we approximate~\eqref{eq: traj} with a Neural ODE~\cite{chen2018neural}, where $f_\theta$ is modeled by a neural network with parameters $\theta$. Denote by $\psi_\theta: \real^k \times \mathcal{T} \to \real^{d|\mathcal{T}|}$ the function that represents the Neural ODE, where $\mathcal{T}$ is a set of time indices. We define the predicted sets $\hX_1,\dotsc,\hX_{T-1} = \psi_\theta(\X_0,\{1,\dotsc,T-1\})$; given an initial set $\X_0$ it returns the approximation of $\eqref{eq: traj}$ for all $t\in\mathcal{T}$. The resulting discrete distributions are $\hat{\mmu}_i := (1/n_i)\sum_j\delta_{x_j}$ for $x_j\in\hX_i$. To match the marginals, we used two approaches of training; \emph{local} or \emph{global}. For the local method, we only predict the next sample, hence given $\X_t$ as initial condition, we predict $\hX_{t+1} = \psi_\theta(\X_t,t+1)$. Whereas for the global, we use the entire trajectories given the initial condition $\X_0$. For both cases, we formulate a loss $L_m$ on the marginals, representing the second term in \eqref{eq: W2_in_thm}. To take into account the first term in \eqref{eq: W2_in_thm}, we add the loss $L_e$, where the integral can be approximate with the forward pass of the ODE solver, and $\lambda_e\geq 0$ is a hyperparameter.
\begin{equation}\label{eq: loss_m_loss_e}
    L_m := \sum_{i=1}^{T-1} W_2(\hat{\mmu}_i,\mmu_i) \qquad L_e :=\lambda_e\sum_{i=0}^{T-2}\int_{i}^{i+1}\|f_\theta(x_t,t)\|_2^2dt
\end{equation}
In practice, to compute the Wasserstein between discrete distributions, we use the implementation from the library Python Optimal Transport~\cite{flamary2021pot}. This is in contrasts with the maximum-likelihood approach in CNF methods which requires evaluation of the instantaneous change of variables at every integration timestep at $O(k^2)$ additional cost per function evaluation~\cite{chen2018neural}. 

We add a final density loss $L_d$ inspired by \cite{tong_trajectorynet_2020} to encourage the trajectories to stay on the manifold. Intuitively, for a predicted point $x\in\hX_t$, we minimize the distance to its k-nearest neighbors in $\X_t$ given a lower bound $h>0$
\begin{equation*}
    L_d:= \lambda_d\sum_{t=1}^{T-1}\sum_{x\in\hX_t}\ell_d(x,t), \,\text{ where }\, \ell_d(x,t) := \sum_{i=1}^k \max(0,\text{min-k}(\{\|x-y\|:y\in\X_t\})-h).
\end{equation*}
We describe the overall training procedure of MIOFlow in algorithm~\ref{alg: mioflow}.

\paragraph{Modeling Diffusion} We add a diffusion term in the trajectories, resulting in the SDE $dX_t = f(X_t,t) dt + \sqrt{\sigma_t}dB_t$ In practice, we learn $T$ parameters $\{\sigma_t\}_{t=0}^{T-1}$ It has the advantage of mapping the same cell to different trajectories stochastically as in many naturalistic systems.  Early in the training process, we notice that it promotes bifurcation, as it helps to spread the initial $\X_0$. As training continues, we empirically observed that $\sigma\to 0$. Hence, as shown in \cite{mikami2004monge}, converging to the $W_2$ formulation. For small $\sigma_t$, we train with a regular ODE solver, essentially corresponding to the Euler–Maruyama method. To model complex diffusion, one would need an SDE solver, thus drastically increasing the computational cost.

\begin{algorithm}
\caption{MIOFlow}
\label{alg: mioflow}
    \begin{algorithmic}
    \State {\bfseries Input:} Datasets $\X_0, \dotsc, \X_{T-1}$, graph kernel $\mK_\epsilon$, maximum scale $K$, noise scale $\xi$, batch size $N$, maximum iteration $n_{max}$, initialized encoder $\phi$ and decoder $\phi^{-1}$, maximum local iteration $n_{local}$, maximum global iteration $n_{global}$, initialized neural ODE $\psi_\theta$.
    \State {\bfseries Output:} Trained neural ODE $\psi_\theta$.
    \State {\bfseries GAE:} $\phi, \phi^{-1} \leftarrow \text{GAE}(\X, \mK_\epsilon, K, \xi, N, n_{max},\phi, \phi^{-1}) $ \Comment{See algorithm~\ref{alg: gae} in SM}
    \If{use GAE} $\X_0,\dotsc, \X_{T-1} \leftarrow \phi(\X_0),\dotsc, \phi(\X_{T-1})$
    \EndIf
    \For{i=1 {\bfseries to } $n_{local}$} \Comment{Local training}
        \For{t=0 {\bfseries to } $T-2$}
           \State $\hX_{t+1} \leftarrow \psi_\theta(\X_t,t+1)$
        \EndFor
        \State $L\leftarrow L_m + L_e + L_d$
        \State update $\psi_\theta$ with gradient descent w.r.t. the loss $L$
    \EndFor
    
    \For{i=1 {\bfseries to } $n_{global}$} \Comment{Global training}
         \State $\hX_{1},\dotsc, \hX_{T-1} \leftarrow \psi_\theta(\X_0,\{1,\dotsc,T-1\})$
        \State $L\leftarrow L_m + L_e + L_d$
        \State update $\psi_\theta$ with gradient descent w.r.t. the loss $L$
    \EndFor
    \end{algorithmic}
\end{algorithm}




\section{Results}\label{sec: results} \vspace{-3mm}

Here, we validate our method empirically on both synthetic test cases and single-cell RNA sequencing datasets chosen to test the capability of MIOFlow to follow divergent trajectories from population snapshot data. Using these test cases we assess whether MIOFlow-derived trajectories qualitatively follow the data manifold efficiently, we also quantitatively assess their accuracy using the Wasserstein $W_1$, and Maximum Mean Discrepancy (MMD)~\cite{gretton2008kernel} on held-out timepoints. The MMD is evaluated with the Gaussian~(G) kernel and the mean~(M). We compare our results to those other methods that preform population flows including TrajectoryNet~\cite{tong_trajectorynet_2020} which is based on a CNF that is regularized to achieve efficient paths, and Diffusion Schr\"odinger's Bridge (DSB)~\cite{de2021diffusion} which is an optimal transport framework for generative modeling. The baseline measure in the quantitative results corresponds to the average distance between the previous and next timepoints for a given time $t$. Additional experiments, such as ablation studies, and details are provided in the supplementary material.


\begin{figure}[th]
    \centering
    \includegraphics[width=0.9\linewidth]{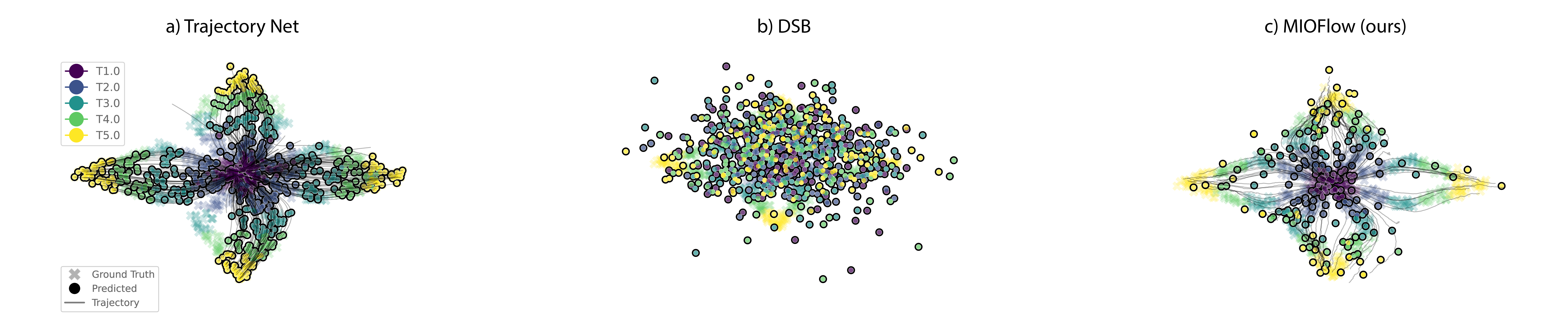}
    \caption{Comparisons between predicted flows of (a) TrajectoryNet, (b) DSB, and (c) MIOFlow on a toy Petal dataset. MIOFlow trajectories both match the data and transition along the data manifold whereas TrajectoryNet takes shortcut paths between the the curved paths and Schroedinger's bridge does not follow the data. Time is indicated by color. Ground truth points are designated by translucent ``X''s in the background, whereas predicted points are opaque circles, whilst trajectories are solid lines.}
    \label{fig:petal_comparison}
\end{figure}

\subsection{Artificial Data} 
\paragraph{Petal}The petal dataset is a simple yet challenging, as it mimics natural dynamics arising in cellular differentiation, including bifurcations and merges. The dynamics start in the center (purple) and evolves through the end of the leaf (yellow). In Fig.~\ref{fig:petal_comparison}, we present the trajectories on the petal dataset for TrajectoryNet, DSB, and MIOFlow. Only our method is able to learn the dynamic, and keeping trajectories on the manifold. We then hold-out timepoint $2$ corresponding to the initial bifurcation, and, in Tab.~\ref{tab:quant}, we compare the interpolation between the three methods. MIOFlow results in a more accurate entropic interpolation for both $W_1$ and MMD(G), while requiring much less computational time. In the supplementary material, we present additional results for DSB on the petal dataset.



\begin{figure}[th]
    \centering
    \includegraphics[width=0.9\linewidth]{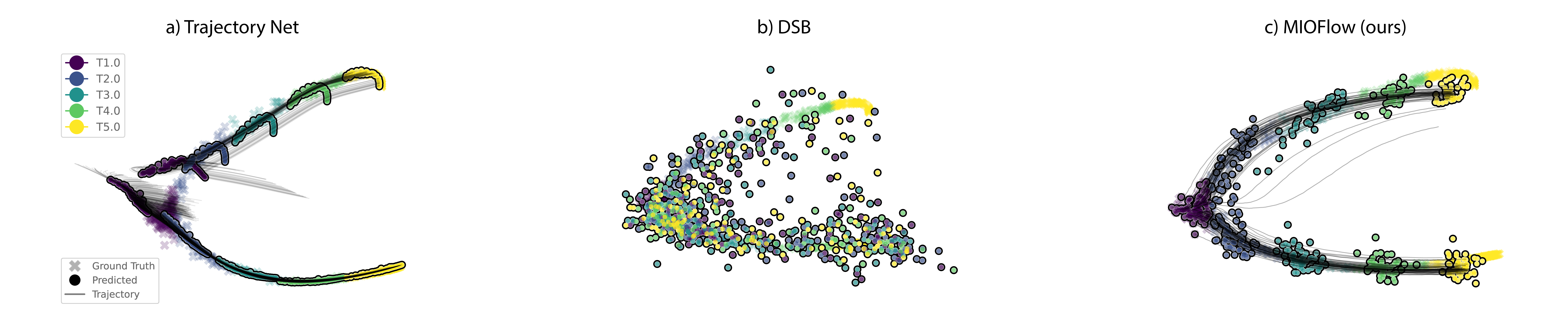}
    \caption{Comparisons between predicted flows of (a) TrajectoryNet, (b) DSB, and (c) MIOFlow on Dyngen~\cite{cannoodt2021spearheading} simulated dataset. MIOFlow trajectories both match the data and transition along the data manifold throughout the timepoints. Time is indicated by color. Ground truth points are designated by translucent ``X''s in the background, whereas predicted points are opaque circles, whilst trajectories are solid lines.}
    \label{fig:dyngen_comparison}
    \vspace{-2mm}
\end{figure}

\paragraph{Dyngen} We use Dyngen~\cite{cannoodt2021spearheading} to simulate a scRNA-seq dataset from a dynamical cellular process. We first embed the observations in five dimensions using the non-linear dimensionality reduction method PHATE~\cite{moon_visualizing_2019}. We discretize the space in five bins, each representing a differentiation stage. This dataset includes one bifurcation. However, it proves more challenging than the petal dataset. The number of samples per bins is not equal, which tends to give more influence to the lower branch. Moreover, the bifurcation is asymmetric; the curve in the lower branch is less pronounce than the upper one, making it harder to learn a continuous bifurcation. In Fig.~\ref{fig:dyngen_comparison}, we show the trajectories for TrajectoryNet, DSB, and our method. MIOFlow is more accurate as it learns continuous bifurcating trajectories along the two main branches. In Tab.~\ref{tab:quant}, we hold-out timepoint 2, and show the interpolation results against the ground truth. MIOFlow is more accurate for all metrics, while requiring much less computational time. We see that in timepoint 1, TrajectoryNet deviates far from the trajectories particularly at the branching point where it struggles with induced heterogeneity (see Fig.~\ref{fig:dyngen_comparison}). We have further quantifications reflecting this in the supplementary material. 

\vspace{-2mm}



\begin{table}[htbp]
  \centering
  \caption{Leave-one-out $W_1$ and MMD with (G)aussian or (M)ean kernel between the predicted and ground truth points as well as training time in (m)inutes and (s)econds on the Petal and Dyngen datasets of TrajectoryNet, DSB, and MIOFlow. Lower is better.}
  \resizebox{\columnwidth}{!}{%
    \begin{tabular}{lcccr|cccr}
    \toprule
          & \multicolumn{1}{l}{Petal} &       &       & \multicolumn{1}{r}{} & \multicolumn{1}{l}{Dyngen} &       &       &  \\
          & $W_1$   & MMD(G) & MMD(M) & Runtime & $W_1$   & MMD(G) & MMD(M) & Runtime \\
    \midrule
    MIOFlow (ours) & \textbf{0.090} & \textbf{0.029} & 0.005 & \textbf{284.60 (s)} & \textbf{0.783} & \textbf{0.199} & \textbf{0.509} & \textbf{95.63} (s) \\
    TrajectoryNet & 0.181 & 0.136 & 0.002 & 64 (m) & 1.499 & 0.303 & 1.806 & 62 (m) \\
    DSB   & 0.199 & 0.157 & 0.008 & 86 (m) & 2.051 & 0.932 & 2.367 & 81 (m) \\
    Baseline & 0.221 & 0.196 & \textbf{$<$0.001} & N/A   & 1.388 & 1.008 & 0.926 & N/A \\
 \bottomrule\end{tabular}%
 \label{tab:quant}
 }
\vspace{-2mm}
\end{table}%

\subsection{Single-Cell Data} 
\paragraph{Embryoid Body}Here we evaluate our method on scRNA-seq data from the dynamic time course of human embryoid body (EB) differentiation over a period of 27 days, measured in 3 timepoints between days 0-3, 6-9, 12-15, 18-21, and 24-27.  We reduce the dimensionality of this dataset to 200  dimensions using PCA. See the supplementary material for further details on the preprocessing steps and the parameters.

\begin{figure}[ht]
    \centering
    \includegraphics[width=0.9
    \linewidth]{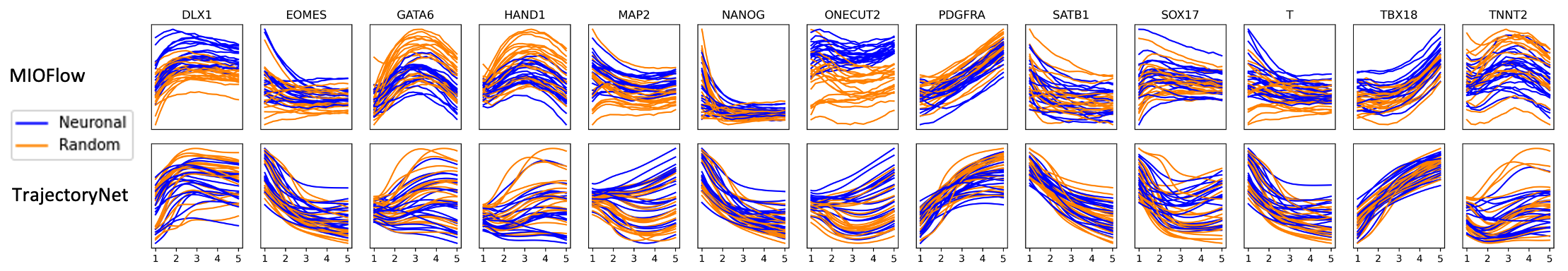}
    \caption{Neuronal gene trajectories vs.\ random trajectories for MIOFlow (top) vs. TrajectoryNet~\cite{tong2021diffusion} (bottom). TrajectoryNet consistently shows a tendency for early divergence in trajectories, examples include ONECUT2 which is biologically known to be higher expressed \cite{moon_visualizing_2019} and HAND2 which is known to peak in neural early progenitors (NEPs). }
    \label{fig:gene_traj}
\end{figure}

By combining PCA and the autoencoder with diffusion geodesic distance described in section~\ref{sec: manifold_reg}, we can decode the trajectories back into the gene space (17846 dimensions), thus describing the evolution of the gene expression of a cell. In Fig.~\ref{fig:gene_traj} we plot the trajectories, at a gene level of 20 neuronal cells against 20 random one. We generally observe distinct differences between TrajectoryNet and MIOFlow. TrajectoryNet (which is shown in the bottom) shows a tendency to have early divergence in trends artificially showing higher heterogeneity in gene trends then exist in neuronal specification processes. Examples include HAND2, which has a non-monotonic trend, first increasing in expression during development from embryonic stem cells to neural early progenitors and then decreasing as the lineages mature \cite{lei2011targeted}. MIOFlow clearly shows this trend whereas TrajectoryNet trends are more chaotic. Another example is ONECUT2 which is consistently known to be higher in cells that diverge towards the neuronal lineage \cite{van2019onecut}. Other examples are discussed in the supplementary material.



\begin{figure}[ht]
\vspace{-2mm}
    \centering
    \includegraphics[width=0.75\linewidth, height=7em]{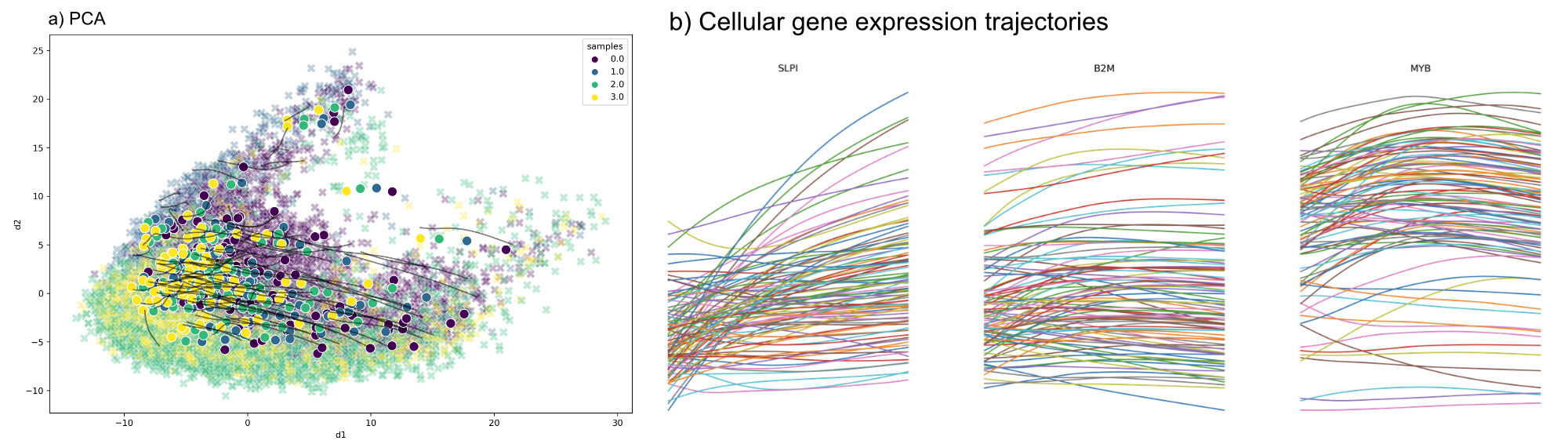} 
    \caption{MIOFlow on the AML dataset. a) PCA Embedding comparing ground truth to MIOFlow predicted points and trajectories. b) Cellular gene expression trajectories for \textit{SLPI}, \textit{B2M}, and \textit{MYB} for MIOFlow.}
    \label{fig:aml-genes}
    \vspace{-4mm}
\end{figure}

In Tab.~\ref{tab:eb_ho_maintext}, we compare the prediction accuracy of MIOFlow whether we use the GAE embedding or no embedding. For the GAE, we show results for two kernels; the Gaussian kernel (scale $0.5$), and the $\alpha$-decay kernel~\cite{moon_visualizing_2019} (based on a 30 nearest neighbors graph). For almost all metrics, the prediction accuracy is greater with the GAE. We present additional results for the choice of kernel in the supplementary material.

\begin{table}[htbp]
  \centering
  \caption{Leave-one-out $(t)$ $W_1$ and MMD with (G)aussian and (M)ean kernel between the predicted and ground truth point on the EB datasets for MIOFlow with the Gaussian, $\alpha$-decay kernel, or without GAE. Lower is better.}
    \begin{tabular}{lcccccc}
    \toprule
    & \multicolumn{3}{c}{$t=2$} & \multicolumn{3}{c}{$t=3$} \\
    \cmidrule{2-7}      & $W_1$   & MMD(G) & MMD(M) & $W_1$   & MMD(G) & MMD(M) \\
    \midrule
    GAE Gaussian & 26.318 & \textbf{0.059} & 101.969 & 32.963 & 0.138 & 223.679\\
    GAE $\alpha$-decay & \textbf{25.744} & 0.061 & \textbf{99.747} & \textbf{32.227} & 0.135 & \textbf{213.465} \\
    No GAE & 29.243 & 0.063 & 126.608 & 35.709 & 0.119 & 246.609 \\
    Baseline       & 33.415 & 0.103 & 227.279 & 35.319 & \textbf{0.095} & 213.492 \\
    \bottomrule
    \end{tabular}%
  \label{tab:eb_ho_maintext}%
\end{table}%

\paragraph{AML Cancer Data with Treatment-based Progression}

We also assessed whether MIOFlow can give insights on treatment resistant cancer cell populations, rather than differentiation. For this purpose, we used data from Fennell et al.~\cite{fennell2022non} in which AF9 mutant cancer cells implanted into the bone marrow of mice as a model of Acute Myeloid Leukemia (AML). The mice are then treated for 7 days with intensive chemotherapy with single cell measurements taken at day 2, day 5, and day 7. Here we show MIOFlow on this dataset along with signatures of leukemic stem cells (LSCs) which are hypothesized to be dominant in chemotherapy resistant cells. We see heterogeneity in the trajectories, the majority of cells surviving to the last timepoint showing an increase in the markers SLPI, MYB, and B2M---all members of the LSC pathway \cite{fennell2022non}. This indicates that not all cells start with this signature but are made to transition to this signature to resist the effects of chemotherapy.

\section{Related Work}
The task of recovering dynamics from multiple snapshots has gained a lot of attention, especially in biology with scRNA-seq data. Static methods such as pseudotime assign an abstract measure of time for all cells depending on an initial state \cite{haghverdi_diffusion_2016,trapnell2014pseudo}. In can be combined to infer the cell lineage, i.e. the principal curves or trajectories~\cite{street2018slingshot}. When estimates of cell growth rates are available, \citet{zhang_optimal_2021} proposed a method based on optimal transport. Another line of research is on the direct rate of differentiation of cell or RNA velocity \cite{la_manno_rna_2018, bergen_rna_2021}. This method does not provide trajectories, but it can be combined to define a transition matrix to infer trajectories~\cite{zhang_inference_2021}. \citet{saelens2019comparison} provide a rigorous comparison of 45 static methods.

Similar methods used the dynamic formulation of optimal transport \cite{bunne_proximal_2022, finlay2020train, hashimoto2016learning,prasad2020optimal,richter2022neural,schiebinger_reconstruction_2017,tong_trajectorynet_2020} to infer trajectories. The dynamics are either modeled with a recurrent neural network \cite{hashimoto2016learning}, a generative adversarial networks \cite{prasad2020optimal}, flows defined by the gradient of an input convex neural network \cite{bunne_proximal_2022}, or a CNF~\cite{tong_trajectorynet_2020}. Here, we circumvent the need for a CNF by modeling the trajectories directly on the support of the distributions, hence maintaining the computational cost low even in high dimension. Some generative methods that interpolate between a prior to a target can also be used to recover trajectories. Other methods use the Schr\"odinger bridge (SB) ~\cite{de2021diffusion,koshizuka2022neural,vargas_solving_2021,wang2021deep} problem to define trajectories between distributions. For example \cite{de2021diffusion}, solve the SB between two distributions using a series of Sinkhorn iterations. However, it does not explicitly learn dynamics, making their interpolations inaccurate.

\section{Conclusion}

We proposed MIOFlow a computationally efficient method to recover trajectories from multiple cross-sectional samples of a time-evolving population. Our method includes a novel manifold learning framework, where we learn an embedding preserving the geodesic distance. We show theoretically how MIOFlow is related to dynamical optimal transport on an underlying manifold, and it's capacity to model stochastic trajectories. On artificial datasets, MIOFlow produced continuous trajectories more in line with the manifold compared to DSB and TrajectoryNet. On scRNA-seq data, we leveraged an autoencoder with the geodesic embedding to learn the trajectories in a lower dimensional space while preserving meaningful distances which can then be decoded in gene space. This way we could visualize the evolution of the gene expression of a cell. 

\section{Limitations and Broader Impact}
Our method does have some limitations. Similarly to other works based on continuous time networks, our work makes use of generic ODE solvers and adjoint dynamics which can fail under stiff dynamics. Like other methods that learn dynamics it is possible that the learned dynamics may become stiff through training leading to decreased model accuracy. In addition, our work is sensitive to the time-bins, and it worked best for relatively uniform sampling. This second issue could potentially be alleviated by using unbalanced optimal transport, which allows transporting distributions with unequal masses. Our work can help understand dynamical systems, and in particular cellular differentiation, and thus to the best of our knowledge presents no potential negative societal impacts.

\begin{ack}
This work was partially funded and supported by CIFAR AI Chair [G.W.], NSERC Discovery
grant 03267 [G.W.], NIH grants 1F30AI157270-01 [M.K.], R01GM135929 [G.W.,
S.K.], R01HD100035, and R01GM130847 [G.W.,
S.K.], NSF Career grant 2047856 [S.K.], the Chan-Zuckerberg Initiative grants CZF2019-182702 and CZF2019-002440 [S.K.], and the Sloan Fellowship FG-2021-15883 [S.K.]. The content provided here is solely the responsibility of the authors and does not necessarily represent the official views of the funding agencies. The funders had no role in study design, data
collection \& analysis, decision to publish, or preparation of the manuscript.
\end{ack}

\bibliographystyle{plainnat}
{\small
\bibliography{tidy}

\begin{thebibliography}{6}
\providecommand{\natexlab}[1]{#1}
\providecommand{\url}[1]{\texttt{#1}}
\expandafter\ifx\csname urlstyle\endcsname\relax
  \providecommand{\doi}[1]{doi: #1}\else
  \providecommand{\doi}{doi: \begingroup \urlstyle{rm}\Url}\fi

\bibitem[Cuturi(2013)]{cuturi_sinkhorn_2013}
Marco Cuturi.
\newblock Sinkhorn {Distances}: {Lightspeed} {Computation} of {Optimal}
  {Transport}.
\newblock In \emph{Advances in {Neural} {Information} {Processing} {Systems}},
  volume~26. Curran Associates, Inc., 2013.

\bibitem[Kidger et~al.(2021)Kidger, Foster, Li, Oberhauser, and
  Lyons]{kidger2021neuralsde}
Patrick Kidger, James Foster, Xuechen Li, Harald Oberhauser, and Terry Lyons.
\newblock Neural {SDE}s as {I}nfinite-{D}imensional {GAN}s.
\newblock \emph{International Conference on Machine Learning}, 2021.

\bibitem[Li et~al.(2020)Li, Wong, Chen, and Duvenaud]{li2020scalable}
Xuechen Li, Ting-Kam~Leonard Wong, Ricky T.~Q. Chen, and David Duvenaud.
\newblock Scalable gradients for stochastic differential equations.
\newblock \emph{International Conference on Artificial Intelligence and
  Statistics}, 2020.

\bibitem[Loshchilov and Hutter(2018)]{loshchilov2018decoupled}
Ilya Loshchilov and Frank Hutter.
\newblock Decoupled weight decay regularization.
\newblock In \emph{International Conference on Learning Representations}, 2018.

\bibitem[Sinkhorn(1964)]{sinkhorn1964relationship}
Richard Sinkhorn.
\newblock A relationship between arbitrary positive matrices and doubly
  stochastic matrices.
\newblock \emph{The annals of mathematical statistics}, 35\penalty0
  (2):\penalty0 876--879, 1964.

\bibitem[Wang(1997)]{wang1997sharp}
Feng-Yu Wang.
\newblock Sharp explicit lower bounds of heat kernels.
\newblock \emph{The Annals of Probability}, 25\penalty0 (4):\penalty0
  1995--2006, 1997.

\end{thebibliography}


\begin{thebibliography}{53}
\providecommand{\natexlab}[1]{#1}
\providecommand{\url}[1]{\texttt{#1}}
\expandafter\ifx\csname urlstyle\endcsname\relax
  \providecommand{\doi}[1]{doi: #1}\else
  \providecommand{\doi}{doi: \begingroup \urlstyle{rm}\Url}\fi

\bibitem[Agrachev and Lee(2010)]{agrachev2010continuity}
Andrei Agrachev and Paul~WY Lee.
\newblock Continuity of optimal control costs and its application to weak kam
  theory.
\newblock \emph{Calculus of Variations and Partial Differential Equations},
  39\penalty0 (1):\penalty0 213--232, 2010.

\bibitem[Belkin and Niyogi(2004)]{belkin_semi-supervised_2004}
Mikhail Belkin and Partha Niyogi.
\newblock Semi-{Supervised} {Learning} on {Riemannian} {Manifolds}.
\newblock \emph{Machine Learning}, 56\penalty0 (1-3):\penalty0 209--239, 2004.

\bibitem[Benamou and Brenier(2000)]{benamou_computational_2000}
Jean-David Benamou and Yann Brenier.
\newblock A computational fluid mechanics solution to the {Monge}-{Kantorovich}
  mass transfer problem.
\newblock \emph{Numerische Mathematik}, 84\penalty0 (3):\penalty0 375--393,
  2000.

\bibitem[Bergen et~al.(2021)Bergen, Soldatov, Kharchenko, and
  Theis]{bergen_rna_2021}
Volker Bergen, Ruslan~A. Soldatov, Peter~V. Kharchenko, and Fabian~J. Theis.
\newblock {RNA} velocity—current challenges and future perspectives.
\newblock \emph{Molecular Systems Biology}, 17\penalty0 (8):\penalty0 e10282,
  2021.

\bibitem[Bunne et~al.(2022)Bunne, {Meng-Papaxanthos}, Krause, and
  Cuturi]{bunne_proximal_2022}
Charlotte Bunne, Laetitia {Meng-Papaxanthos}, Andreas Krause, and Marco Cuturi.
\newblock Proximal {{Optimal Transport Modeling}} of {{Population Dynamics}}.
\newblock \emph{AISTATS}, 2022.

\bibitem[Cannoodt et~al.(2021)Cannoodt, Saelens, Deconinck, and
  Saeys]{cannoodt2021spearheading}
Robrecht Cannoodt, Wouter Saelens, Louise Deconinck, and Yvan Saeys.
\newblock Spearheading future omics analyses using dyngen, a multi-modal
  simulator of single cells.
\newblock \emph{Nature Communications}, 12\penalty0 (1):\penalty0 1--9, 2021.

\bibitem[Chen et~al.(2018)Chen, Rubanova, Bettencourt, and
  Duvenaud]{chen2018neural}
Ricky~TQ Chen, Yulia Rubanova, Jesse Bettencourt, and David~K Duvenaud.
\newblock Neural ordinary differential equations.
\newblock \emph{Advances in Neural Information Processing Systems}, 31, 2018.

\bibitem[Cheng and Wu(2022)]{cheng2022eigen}
Xiuyuan Cheng and Nan Wu.
\newblock Eigen-convergence of gaussian kernelized graph laplacian by manifold
  heat interpolation.
\newblock \emph{Applied and Computational Harmonic Analysis}, 61:\penalty0
  132--190, 2022.

\bibitem[Coifman and Lafon(2006)]{coifman_diffusion_2006}
Ronald~R. Coifman and Stéphane Lafon.
\newblock Diffusion maps.
\newblock \emph{Applied and Computational Harmonic Analysis}, 21\penalty0
  (1):\penalty0 5--30, 2006.

\bibitem[De~Bortoli et~al.(2021)De~Bortoli, Thornton, Heng, and
  Doucet]{de2021diffusion}
Valentin De~Bortoli, James Thornton, Jeremy Heng, and Arnaud Doucet.
\newblock Diffusion schr{\"o}dinger bridge with applications to score-based
  generative modeling.
\newblock \emph{Advances in Neural Information Processing Systems}, 34, 2021.

\bibitem[Duque et~al.(2020)Duque, Morin, Wolf, and Moon]{duque2020extendable}
Andr{\'e}s~F Duque, Sacha Morin, Guy Wolf, and Kevin Moon.
\newblock Extendable and invertible manifold learning with geometry regularized
  autoencoders.
\newblock In \emph{2020 IEEE International Conference on Big Data (Big Data)},
  pages 5027--5036. IEEE, 2020.

\bibitem[Fennell et~al.(2022)Fennell, Vassiliadis, Lam, Martelotto, Balic,
  Hollizeck, Weber, Semple, Wang, Miles, et~al.]{fennell2022non}
Katie~A Fennell, Dane Vassiliadis, Enid~YN Lam, Luciano~G Martelotto, Jesse~J
  Balic, Sebastian Hollizeck, Tom~S Weber, Timothy Semple, Qing Wang, Denise~C
  Miles, et~al.
\newblock Non-genetic determinants of malignant clonal fitness at single-cell
  resolution.
\newblock \emph{Nature}, 601\penalty0 (7891):\penalty0 125--131, 2022.

\bibitem[Finlay et~al.(2020)Finlay, Jacobsen, Nurbekyan, and
  Oberman]{finlay2020train}
Chris Finlay, J{\"o}rn-Henrik Jacobsen, Levon Nurbekyan, and Adam Oberman.
\newblock How to train your neural ode: the world of jacobian and kinetic
  regularization.
\newblock In \emph{International conference on machine learning}, pages
  3154--3164. PMLR, 2020.

\bibitem[Flamary et~al.(2021)Flamary, Courty, Gramfort, Alaya, Boisbunon,
  Chambon, Chapel, Corenflos, Fatras, Fournier, Gautheron, Gayraud, Janati,
  Rakotomamonjy, Redko, Rolet, Schutz, Seguy, Sutherland, Tavenard, Tong, and
  Vayer]{flamary2021pot}
R{\'e}mi Flamary, Nicolas Courty, Alexandre Gramfort, Mokhtar~Z. Alaya,
  Aur{\'e}lie Boisbunon, Stanislas Chambon, Laetitia Chapel, Adrien Corenflos,
  Kilian Fatras, Nemo Fournier, L{\'e}o Gautheron, Nathalie~T.H. Gayraud,
  Hicham Janati, Alain Rakotomamonjy, Ievgen Redko, Antoine Rolet, Antony
  Schutz, Vivien Seguy, Danica~J. Sutherland, Romain Tavenard, Alexander Tong,
  and Titouan Vayer.
\newblock Pot: Python optimal transport.
\newblock \emph{Journal of Machine Learning Research}, 22\penalty0
  (78):\penalty0 1--8, 2021.

\bibitem[Gardiner et~al.(1985)]{gardiner1985handbook}
Crispin~W Gardiner et~al.
\newblock \emph{Handbook of stochastic methods}, volume~3.
\newblock springer Berlin, 1985.

\bibitem[Ghoussoub et~al.(2021)Ghoussoub, Kim, and
  Palmer]{ghoussoub2021solution}
Nassif Ghoussoub, Young-Heon Kim, and Aaron~Zeff Palmer.
\newblock A solution to the monge transport problem for brownian martingales.
\newblock \emph{the Annals of Probability}, 49\penalty0 (2):\penalty0 877--907,
  2021.

\bibitem[Grathwohl et~al.(2019)Grathwohl, Chen, Bettencourt, Sutskever, and
  Duvenaud]{grathwohl_ffjord:_2019}
Will Grathwohl, Ricky T.~Q. Chen, Jesse Bettencourt, Ilya Sutskever, and David
  Duvenaud.
\newblock {{FFJORD}}: {{Free-form Continuous Dynamics}} for {{Scalable
  Reversible Generative Models}}.
\newblock In \emph{7th {{International Conference}} on {{Learning
  Representations}}}, 2019.

\bibitem[Gretton et~al.(2008)Gretton, Borgwardt, Rasch, Sch{\"o}lkopf, and
  Smola]{gretton2008kernel}
Arthur Gretton, Karsten~M Borgwardt, Malte~J Rasch, Bernhard Sch{\"o}lkopf, and
  Alexander Smola.
\newblock A kernel method for the two-sample problem.
\newblock \emph{Journal of Machine Learning Research}, 1:\penalty0 1--10, 2008.

\bibitem[Haghverdi et~al.(2016)Haghverdi, Büttner, Wolf, Buettner, and
  Theis]{haghverdi_diffusion_2016}
Laleh Haghverdi, Maren Büttner, F.~Alexander Wolf, Florian Buettner, and
  Fabian~J. Theis.
\newblock Diffusion pseudotime robustly reconstructs lineage branching.
\newblock \emph{Nature Methods}, 13\penalty0 (10):\penalty0 845--848, 2016.

\bibitem[Hashimoto et~al.(2016)Hashimoto, Gifford, and
  Jaakkola]{hashimoto2016learning}
Tatsunori Hashimoto, David Gifford, and Tommi Jaakkola.
\newblock Learning population-level diffusions with generative rnns.
\newblock In \emph{International Conference on Machine Learning}, pages
  2417--2426. PMLR, 2016.

\bibitem[Ho et~al.(2020)Ho, Jain, and Abbeel]{ho_denoising_2020}
Jonathan Ho, Ajay Jain, and Pieter Abbeel.
\newblock Denoising {{Diffusion Probabilistic Models}}.
\newblock \emph{Advances in Neural Information Processing Systems},
  33:\penalty0 6840--6851, 2020.

\bibitem[Koshizuka and Sato(2022)]{koshizuka2022neural}
Takeshi Koshizuka and Issei Sato.
\newblock Neural lagrangian schr\"odinger bridge.
\newblock \emph{arXiv preprint arXiv:2204.04853}, 2022.

\bibitem[La~Manno et~al.(2018)La~Manno, Soldatov, Zeisel, Braun, Hochgerner,
  Petukhov, Lidschreiber, Kastriti, Lönnerberg, Furlan, Fan, Borm, Liu, van
  Bruggen, Guo, He, Barker, Sundström, Castelo-Branco, Cramer, Adameyko,
  Linnarsson, and Kharchenko]{la_manno_rna_2018}
Gioele La~Manno, Ruslan Soldatov, Amit Zeisel, Emelie Braun, Hannah Hochgerner,
  Viktor Petukhov, Katja Lidschreiber, Maria~E. Kastriti, Peter Lönnerberg,
  Alessandro Furlan, Jean Fan, Lars~E. Borm, Zehua Liu, David van Bruggen,
  Jimin Guo, Xiaoling He, Roger Barker, Erik Sundström, Gonçalo
  Castelo-Branco, Patrick Cramer, Igor Adameyko, Sten Linnarsson, and Peter~V.
  Kharchenko.
\newblock {RNA} velocity of single cells.
\newblock \emph{Nature}, 560\penalty0 (7719):\penalty0 494--498, 2018.

\bibitem[Lasry and Lions(2007)]{lasry2007mean}
Jean-Michel Lasry and Pierre-Louis Lions.
\newblock Mean field games.
\newblock \emph{Japanese journal of mathematics}, 2\penalty0 (1):\penalty0
  229--260, 2007.

\bibitem[Leeb and Coifman(2016)]{leeb_holderlipschitz_2016}
William Leeb and Ronald Coifman.
\newblock Hölder–{Lipschitz} {Norms} and {Their} {Duals} on {Spaces} with
  {Semigroups}, with {Applications} to {Earth} {Mover}’s {Distance}.
\newblock \emph{J Fourier Anal Appl}, 22\penalty0 (4):\penalty0 910--953, 2016.

\bibitem[Lei and Howard(2011)]{lei2011targeted}
Jun Lei and Marthe~J Howard.
\newblock Targeted deletion of hand2 in enteric neural precursor cells affects
  its functions in neurogenesis, neurotransmitter specification and
  gangliogenesis, causing functional aganglionosis.
\newblock \emph{Development}, 138\penalty0 (21):\penalty0 4789--4800, 2011.

\bibitem[Mathieu and Nickel(2020)]{mathieu2020riemannian}
Emile Mathieu and Maximilian Nickel.
\newblock Riemannian continuous normalizing flows.
\newblock \emph{Advances in Neural Information Processing Systems},
  33:\penalty0 2503--2515, 2020.

\bibitem[Mikami(2004)]{mikami2004monge}
Toshio Mikami.
\newblock Monge’s problem with a quadratic cost by the zero-noise limit of
  h-path processes.
\newblock \emph{Probability theory and related fields}, 129\penalty0
  (2):\penalty0 245--260, 2004.

\bibitem[Mikami and Thieullen(2006)]{mikami2006duality}
Toshio Mikami and Mich{\`e}le Thieullen.
\newblock Duality theorem for the stochastic optimal control problem.
\newblock \emph{Stochastic processes and their applications}, 116\penalty0
  (12):\penalty0 1815--1835, 2006.

\bibitem[Mishne et~al.(2019)Mishne, Shaham, Cloninger, and
  Cohen]{mishne2019diffusion}
Gal Mishne, Uri Shaham, Alexander Cloninger, and Israel Cohen.
\newblock Diffusion nets.
\newblock \emph{Applied and Computational Harmonic Analysis}, 47\penalty0
  (2):\penalty0 259--285, 2019.

\bibitem[Moon et~al.(2018)Moon, Stanley, Burkhardt, {van Dijk}, Wolf, and
  Krishnaswamy]{moon_manifold_2018}
Kevin~R. Moon, Jay~S. Stanley, Daniel Burkhardt, David {van Dijk}, Guy Wolf,
  and Smita Krishnaswamy.
\newblock Manifold learning-based methods for analyzing single-cell
  {{RNA-sequencing}} data.
\newblock \emph{Current Opinion in Systems Biology}, 7:\penalty0 36--46, 2018.

\bibitem[Moon et~al.(2019)Moon, van Dijk, Wang, Gigante, Burkhardt, Chen, Yim,
  Elzen, Hirn, Coifman, Ivanova, Wolf, and Krishnaswamy]{moon_visualizing_2019}
Kevin~R. Moon, David van Dijk, Zheng Wang, Scott Gigante, Daniel~B. Burkhardt,
  William~S. Chen, Kristina Yim, Antonia van~den Elzen, Matthew~J. Hirn,
  Ronald~R. Coifman, Natalia~B. Ivanova, Guy Wolf, and Smita Krishnaswamy.
\newblock Visualizing structure and transitions in high-dimensional biological
  data.
\newblock \emph{Nat Biotechnol}, 37\penalty0 (12):\penalty0 1482--1492, 2019.

\bibitem[Pavon et~al.(2021)Pavon, Trigila, and Tabak]{pavon2021data}
Michele Pavon, Giulio Trigila, and Esteban~G Tabak.
\newblock The data-driven schr{\"o}dinger bridge.
\newblock \emph{Communications on Pure and Applied Mathematics}, 74\penalty0
  (7):\penalty0 1545--1573, 2021.

\bibitem[Peyré and Cuturi(2020)]{peyre_computational_2020}
Gabriel Peyré and Marco Cuturi.
\newblock Computational {Optimal} {Transport}.
\newblock \emph{arXiv:1803.00567}, 2020.

\bibitem[Prasad et~al.(2020)Prasad, Yang, and Uhler]{prasad2020optimal}
Neha Prasad, Karren Yang, and Caroline Uhler.
\newblock Optimal transport using gans for lineage tracing.
\newblock \emph{arXiv preprint arXiv:2007.12098}, 2020.

\bibitem[Richter-Powell et~al.(2022)Richter-Powell, Lipman, and
  Chen]{richter2022neural}
Jack Richter-Powell, Yaron Lipman, and Ricky~TQ Chen.
\newblock Neural conservation laws: A divergence-free perspective.
\newblock \emph{arXiv preprint arXiv:2210.01741}, 2022.

\bibitem[Saelens et~al.(2019)Saelens, Cannoodt, Todorov, and
  Saeys]{saelens2019comparison}
Wouter Saelens, Robrecht Cannoodt, Helena Todorov, and Yvan Saeys.
\newblock A comparison of single-cell trajectory inference methods.
\newblock \emph{Nature biotechnology}, 37\penalty0 (5):\penalty0 547--554,
  2019.

\bibitem[Schiebinger et~al.(2019)Schiebinger, Shu, Tabaka, Cleary, Subramanian,
  Solomon, Liu, Lin, Berube, Lee, Chen, Brumbaugh, Rigollet, Hochedlinger,
  Jaenisch, Regev, and Lander]{schiebinger_reconstruction_2017}
Geoffrey Schiebinger, Jian Shu, Marcin Tabaka, Brian Cleary, Vidya Subramanian,
  Aryeh Solomon, Siyan Liu, Stacie Lin, Peter Berube, Lia Lee, Jenny Chen,
  Justin Brumbaugh, Philippe Rigollet, Konrad Hochedlinger, Rudolf Jaenisch,
  Aviv Regev, and Eric~S. Lander.
\newblock Reconstruction of developmental landscapes by optimal-transport
  analysis of single-cell gene expression sheds light on cellular
  reprogramming.
\newblock \emph{Cell}, 2019.

\bibitem[Singer(2006)]{singer_graph_2006}
A.~Singer.
\newblock From graph to manifold {Laplacian}: {The} convergence rate.
\newblock \emph{Applied and Computational Harmonic Analysis}, 21\penalty0
  (1):\penalty0 128--134, 2006.

\bibitem[Solomon et~al.(2015)Solomon, De~Goes, Peyr{\'e}, Cuturi, Butscher,
  Nguyen, Du, and Guibas]{solomon2015convolutional}
Justin Solomon, Fernando De~Goes, Gabriel Peyr{\'e}, Marco Cuturi, Adrian
  Butscher, Andy Nguyen, Tao Du, and Leonidas Guibas.
\newblock Convolutional wasserstein distances: Efficient optimal transportation
  on geometric domains.
\newblock \emph{ACM Transactions on Graphics (ToG)}, 34\penalty0 (4):\penalty0
  1--11, 2015.

\bibitem[Song and Ermon(2019)]{song_generative_2019}
Yang Song and Stefano Ermon.
\newblock Generative {{Modeling}} by {{Estimating Gradients}} of the {{Data
  Distribution}}.
\newblock \emph{Advances in Neural Information Processing Systems}, pages
  11918--11930, 2019.

\bibitem[Song et~al.(2020)Song, Sohl-Dickstein, Kingma, Kumar, Ermon, and
  Poole]{song2020score}
Yang Song, Jascha Sohl-Dickstein, Diederik~P Kingma, Abhishek Kumar, Stefano
  Ermon, and Ben Poole.
\newblock Score-based generative modeling through stochastic differential
  equations.
\newblock In \emph{International Conference on Learning Representations}, 2020.

\bibitem[Street et~al.(2018)Street, Risso, Fletcher, Das, Ngai, Yosef, Purdom,
  and Dudoit]{street2018slingshot}
Kelly Street, Davide Risso, Russell~B Fletcher, Diya Das, John Ngai, Nir Yosef,
  Elizabeth Purdom, and Sandrine Dudoit.
\newblock Slingshot: cell lineage and pseudotime inference for single-cell
  transcriptomics.
\newblock \emph{BMC genomics}, 19\penalty0 (1):\penalty0 1--16, 2018.

\bibitem[Tong et~al.(2020)Tong, Huang, Wolf, Dijk, and
  Krishnaswamy]{tong_trajectorynet_2020}
Alexander Tong, Jessie Huang, Guy Wolf, David~Van Dijk, and Smita Krishnaswamy.
\newblock {TrajectoryNet}: {A} {Dynamic} {Optimal} {Transport} {Network} for
  {Modeling} {Cellular} {Dynamics}.
\newblock In \emph{Proceedings of the 37th {International} {Conference} on
  {Machine} {Learning}}, pages 9526--9536. PMLR, 2020.

\bibitem[Tong et~al.(2022)Tong, Huguet, Shung, Natik, Kuchroo, Lajoie, Wolf,
  and Krishnaswamy]{tong2022embedding}
Alexander Tong, Guillaume Huguet, Dennis Shung, Amine Natik, Manik Kuchroo,
  Guillaume Lajoie, Guy Wolf, and Smita Krishnaswamy.
\newblock Embedding signals on graphs with unbalanced diffusion earth mover’s
  distance.
\newblock In \emph{ICASSP 2022-2022 IEEE International Conference on Acoustics,
  Speech and Signal Processing (ICASSP)}, pages 5647--5651, 2022.

\bibitem[Tong et~al.(2021)Tong, Huguet, Natik, MacDonald, Kuchroo, Coifman,
  Wolf, and Krishnaswamy]{tong2021diffusion}
Alexander~Y Tong, Guillaume Huguet, Amine Natik, Kincaid MacDonald, Manik
  Kuchroo, Ronald Coifman, Guy Wolf, and Smita Krishnaswamy.
\newblock Diffusion earth mover’s distance and distribution embeddings.
\newblock In \emph{International Conference on Machine Learning}, pages
  10336--10346. PMLR, 2021.

\bibitem[Trapnell et~al.(2014)Trapnell, Cacchiarelli, Grimsby, Pokharel, Li,
  Morse, Lennon, Livak, Mikkelsen, and Rinn]{trapnell2014pseudo}
Cole Trapnell, Davide Cacchiarelli, Jonna Grimsby, Prapti Pokharel, Shuqiang
  Li, Michael Morse, Niall~J Lennon, Kenneth~J Livak, Tarjei~S Mikkelsen, and
  John~L Rinn.
\newblock Pseudo-temporal ordering of individual cells reveals dynamics and
  regulators of cell fate decisions.
\newblock \emph{Nature biotechnology}, 32\penalty0 (4):\penalty0 381, 2014.

\bibitem[van~der Raadt et~al.(2019)van~der Raadt, van Gestel, Nadif~Kasri, and
  Albers]{van2019onecut}
Jori van~der Raadt, Sebastianus~HC van Gestel, Nael Nadif~Kasri, and Cornelis~A
  Albers.
\newblock Onecut transcription factors induce neuronal characteristics and
  remodel chromatin accessibility.
\newblock \emph{Nucleic acids research}, 47\penalty0 (11):\penalty0 5587--5602,
  2019.

\bibitem[Vargas et~al.(2021)Vargas, Thodoroff, Lamacraft, and
  Lawrence]{vargas_solving_2021}
Francisco Vargas, Pierre Thodoroff, Austen Lamacraft, and Neil Lawrence.
\newblock Solving {Schrödinger} {Bridges} via {Maximum} {Likelihood}.
\newblock \emph{Entropy}, 23\penalty0 (9):\penalty0 1134, 2021.

\bibitem[Villani(2009)]{villani2009optimal}
C{\'e}dric Villani.
\newblock \emph{Optimal transport: old and new}, volume 338.
\newblock Springer, 2009.

\bibitem[Wang et~al.(2021)Wang, Jiao, Xu, Wang, and Yang]{wang2021deep}
Gefei Wang, Yuling Jiao, Qian Xu, Yang Wang, and Can Yang.
\newblock Deep generative learning via schr{\"o}dinger bridge.
\newblock In \emph{International Conference on Machine Learning}, pages
  10794--10804. PMLR, 2021.

\bibitem[Zhang et~al.(2021)Zhang, Afanassiev, Greenstreet, Matsumoto, and
  Schiebinger]{zhang_optimal_2021}
Stephen Zhang, Anton Afanassiev, Laura Greenstreet, Tetsuya Matsumoto, and
  Geoffrey Schiebinger.
\newblock Optimal transport analysis reveals trajectories in steady-state
  systems.
\newblock Technical report, 2021.

\bibitem[Zhang and Zhang(2021)]{zhang_inference_2021}
Ziqi Zhang and Xiuwei Zhang.
\newblock Inference of high-resolution trajectories in single-cell {RNA}-seq
  data by using {RNA} velocity.
\newblock \emph{Cell Reports Methods}, 1\penalty0 (6):\penalty0 100095, 2021.

\end{thebibliography}
}

\clearpage



\begin{enumerate}

\item For all authors...
\begin{enumerate}
  \item Do the main claims made in the abstract and introduction accurately reflect the paper's contributions and scope? 
    \answerYes{}
  \item Did you describe the limitations of your work?
    \answerYes{In the limitations and broader impact section.}
  \item Did you discuss any potential negative societal impacts of your work?
   
  \answerYes{In the limitations and broader impact section. We do not foresee any immediate negative societal impacts of this work.}
  \item Have you read the ethics review guidelines and ensured that your paper conforms to them?
    \answerYes{}
\end{enumerate}

\item If you are including theoretical results...
\begin{enumerate}
  \item Did you state the full set of assumptions of all theoretical results?
    
    \answerYes{We stated all assumptions in the main body of the text and provide additional details in the supplementary materials.}
\item Did you include complete proofs of all theoretical results?

\answerYes{Complete proofs are available in the supplement.}
\end{enumerate}

\item If you ran experiments...
\begin{enumerate}
  \item Did you include the code, data, and instructions needed to reproduce the main experimental results (either in the supplemental material or as a URL)?
    
    \answerYes{Code and links to publicly available data are provided in the supplemental material.}
  \item Did you specify all the training details (e.g., data splits, hyperparameters, how they were chosen)?
    
    \answerYes{All training details are provided in the supplement.}
        \item Did you report error bars (e.g., with respect to the random seed after running experiments multiple times)?
    \answerYes{}
        \item Did you include the total amount of compute and the type of resources used (e.g., type of GPUs, internal cluster, or cloud provider)?
    
    \answerYes{Compute resources are discussed in the supplement.}
\end{enumerate}

\item If you are using existing assets (e.g., code, data, models) or curating/releasing new assets...
\begin{enumerate}
  \item If your work uses existing assets, did you cite the creators?
    \answerYes{}
  \item Did you mention the license of the assets?
    \answerYes{See supplement}
  \item Did you include any new assets either in the supplemental material or as a URL?
    \answerYes{}
  \item Did you discuss whether and how consent was obtained from people whose data you're using/curating?
    \answerNA{}
  \item Did you discuss whether the data you are using/curating contains personally identifiable information or offensive content?
    
    \answerYes{Our data does not contain any personally identifiable information or offensive content. The Embryoid body data is a cell culture, and the AML Cancer dataset is patient derived but contains no identifiable information.}
\end{enumerate}

\item If you used crowdsourcing or conducted research with human subjects...
\begin{enumerate}
  \item Did you include the full text of instructions given to participants and screenshots, if applicable?
    \answerNA{}
  \item Did you describe any potential participant risks, with links to Institutional Review Board (IRB) approvals, if applicable?
    \answerNA{}
  \item Did you include the estimated hourly wage paid to participants and the total amount spent on participant compensation?
    \answerNA{}
\end{enumerate}

\end{enumerate}


\clearpage
\appendix
\section{Theory and Algorithm}
\subsection{Geodesic Autoencoder Algorithm}
Using the definitions from Sec.~\ref{sec:MIOFlow}, we present the algorithm to train the geodesic autoencoder. 

\begin{algorithm}
\caption{Geodesic Autoencoder (GAE)}\label{alg: gae}
    \begin{algorithmic}
    \State {\bfseries Input:} Dataset $\X$ of size $n$, graph kernel $\mK_\epsilon$, maximum scale $K$, noise scale $\xi$, batch size $N$, maximum iteration $n_{max}$, initialized encoder $\phi$ and decoder $\phi^{-1}$.
    \State {\bfseries Output:} Trained encoder $\phi$ and decoder $\phi^{-1}$.
    \For{i=1 {\bfseries to } $n_{max}$}
        \State Sample batch $\{x_1,\dotsc,x_N\}\subseteq \X$ of size $N$
        \State $\mQ \leftarrow \text{diag}(\sum_j (\mK_\epsilon)_{ij})$
        \State $\mM_\epsilon \leftarrow \mQ^{-1} \mK_\epsilon\mQ^{-1}$
        \State $\mD \leftarrow \text{diag}(\sum_j (\mM_\epsilon)_{ij})$
        \State $\mP_\epsilon \leftarrow \mD^{-1}\mM_\epsilon$ \Comment{$N\times N$ diffusion matrix}
        \State $G_\alpha\leftarrow 2^{-K\alpha}\,\text{pairwise\_distances}((\mP_\epsilon)_{1:},\dotsc,(\mP_\epsilon)_{N:})$ 
         \For{k=1 {\bfseries to } K}
            \State $G_\alpha\leftarrow G_\alpha +  2^{-(K-k)\alpha}\,\text{pairwise\_distance}((\mP^{2^k}_\epsilon)_{1:},\dotsc,(\mP^{2^k}_\epsilon)_{N:})$
         \EndFor
         \State $\Tilde{x}_1,\dotsc,\Tilde{x}_N \leftarrow x_1 + \xi z_1, \dots, x_N + \xi z_N$, where $\{z_1,\dotsc,z_N\}\sim\mathcal{N}(0,1)$
         \State $L \leftarrow$ MSE$\big(\text{pairwise\_distances}(\phi(\tilde{x}_1),\dotsc,\phi(\tilde{x}_N)), G_\alpha\big)$
         \State $L_r\leftarrow \sum_i \|\phi^{-1}\circ\phi(\tilde{x}_i) - x\|_2 $
         \State {\bfseries Update} $\phi, \phi^{-1}$ with gradient descent w.r.t. $L(\phi),L_r(\phi,\phi^{-1})$
    \EndFor
    \end{algorithmic}
\end{algorithm}

\subsection{Proof of Thm.2}

\begin{customthm}{2}
Assuming $\X$ is sampled from a closed Riemannian manifold $\M$ with geodesic $d_\M$, then for a sufficiently large $K,N$ and small $\epsilon>0$, we have $G_\alpha(x_i,x_j) \simeq d_\M^{2\alpha}(x_i,x_j)$, for all $\alpha\in(0,1/2)$. 
\end{customthm}

\begin{proof}
We adapt the convergence proof from~\cite{tong2021diffusion}, it can be summarized in two parts. First, the equivalence of the distances defined by $P_\epsilon^{t/\epsilon}$ and the heat operator $H_t$. This relies on the convergence results $P_\epsilon^{t/\epsilon}\to H_t$ as $\epsilon\to 0$ from \citet{coifman_diffusion_2006}. Second, the approximation of the operator $P_\epsilon$ based on a finite set of points. Throughout the proof we use the same notation $\delta_x$ for the delta Dirac function, and $\delta_{x_i}$ for a row vector with $i$-th element $1$, and $0$ otherwise. 

For a family of operator $(A_t)_{t\in\real}$ we define 
\begin{equation*}
    D_{\alpha,A}(x,y) := \sum_{k\geq 0}2^{-k
    \alpha}||A_{2^{-k}}(x,\cdot)-A_{2^{-k}}(y,\cdot)||_1,
\end{equation*}
and equivalently $D_{\alpha,A}(x,y) = \sum 2^{-k
\alpha}||A_{2^{-k}}(\delta_x-\delta_y)||_1$, where $A\delta_x:=\int A(u,\cdot)\delta_x(u)du$. This definition is the same as Def.~\ref{def:diffusion_alpha}, but for an arbitrary operator. Our final goal is to show that  $G_\alpha(x_i,x_j)\simeq D_{\alpha, H_t}(x_i,x_j)$, and then conclude with Thm.~\ref{thm: geodesic_Leeb} which gives the equivalence $D_{\alpha, H_t}(x_i,x_j) \simeq d^{2\alpha}_{\M}(x_i,x_j)$.

We first want to show that, for a sufficiently small $\epsilon>0$, we have $D_{\alpha,P_\epsilon^{t/\epsilon}}(x,y)\simeq D_{\alpha, H_t}$, where $P$ is the anisotropic operator defined in Sec.~\ref{sec: manifold geodesics}. From \citet{coifman_diffusion_2006}, we have $\|P_\epsilon^{t/\epsilon}-H_t\|_{L^2(\M)}\to 0$ as $\epsilon\to 0$, and also $\|P_\epsilon^{t/\epsilon}-H_t\|_{L^1(\M)}\to 0$. 

Now let $\Gamma_t := P_\epsilon^{t/\epsilon}-H_t$, and for $\gamma>0$ choose $\epsilon>0$ such that $\| \Gamma_t (\delta_x-\delta_y) \|_1<\gamma\|\delta_x-\delta_y \|_1=2\gamma$ for $x\neq y$, we have
\begin{equation*}
    D_{\alpha,\Gamma_t}(x,y) = \sum_{k\geq0} 2^{-k
\alpha}||\Gamma_{2^{-k}}(\delta_x-\delta_y)||_1 \leq  2\gamma \sum_{k\geq0} 2^{-k
\alpha} = \frac{2\gamma}{1-2^{-\alpha}}.
\end{equation*}
Thus, for all $t>0$, and all $\gamma>0$ there exists $\epsilon>0$ such that $ D_{\alpha,\Gamma_t}(x,y)<\gamma$. From the reverse triangle inequality, we have
\begin{equation*}
    |D_{\alpha,P_\epsilon^{t/\epsilon}}(x,y)- D_{\alpha, H_t}|\leq  D_{\alpha,\Gamma_t}(x,y),
\end{equation*} using $D_{\alpha,\Gamma_t}(x,y)<\gamma$, we obtain
\begin{equation*}
     D_{\alpha, H_t}(x,y)-\gamma \leq D_{\alpha,P_\epsilon^{t/\epsilon}}(x,y) \leq  D_{\alpha, H_t}(x,y) +\gamma.
\end{equation*}
According to \citeSupp{wang1997sharp} we can lower bound the heat kernel, and thus the distance $D_{\alpha, H_t}(x,y)>C$ for some $C>0$. For $\gamma<C/2$, and a sufficiently small $\epsilon>0$, we have 
\begin{equation*}
    (1/2)D_{\alpha, H_t}(x,y)\leq  D_{\alpha,P_\epsilon^{t/\epsilon}}(x,y)\leq(3/2)D_{\alpha, H_t}(x,y).
\end{equation*}
This proves our first claim, that $D_{\alpha,P_\epsilon^{t/\epsilon}}(x,y)\simeq D_{\alpha, H_t}$ for small $\epsilon>0$. 

Next, we consider the approximation of the operator $P_\epsilon^{t/\epsilon}$ with a finite set of points. Now we define
\begin{equation*}
    G_{K,\epsilon,\alpha}(x_i,x_j) := \sum_{k=0}^K 2^{-k\alpha} || (\delta_{x_i}-\delta_{x_j})\mP_\epsilon^{2^{-k}/\epsilon} ||_1,
\end{equation*}
if $\epsilon=2^{-K}$, then the first term of $G_\alpha(x_i,x_j)$ is equal to $G_{K,\epsilon,\alpha}(x_i,x_j)$. Since we will let $K\to\infty$, we ignore the second term in $G_\alpha(x_i,x_j)$, as it converges to zero. Similar to \citet{coifman_diffusion_2006} (Sec.~5), we have $\mP\to P$ as $n\to\infty$, using Monte-Carlo integration (approximation of integral with summation by the law of large numbers). By the strong law of large numbers, the convergence is with probability one. For a finite number of samples, we have a high probability bound on the convergence, see for example \cite{cheng2022eigen,singer_graph_2006}. Now we let $N:=\min(K,n)$, and therefore
\begin{equation*}
    \lim_{N\to\infty} G_{K,\epsilon,\alpha}(x_i,x_j) = D_{\alpha,P_\epsilon^{t/\epsilon}}(x_i,x_j).
\end{equation*}
Hence, if we take $\epsilon:=2^{-K}$, we have for sufficiently large $n$ and $K$ (implying small $\epsilon>0$), $G_\alpha(x_i,x_j)\simeq D_{\alpha, H_t} \simeq d^{2\alpha}_{\M}(x_i,x_j)$ from Thm.~\ref{thm: geodesic_Leeb}.
\end{proof}

\subsection{Proof of Thm.3}

\begin{customthm}{3}
We consider a time-varying vector field $f(x,t)$ defining the trajectories $dX_t = f(X_t,t)dt$ with density $\rho_t$, and a dissimilarity between distributions such that $D(\mu,\nu)=0$ iff $\mu=\nu$. Given these assumptions, there exist a sufficiently large $\lambda>0$ such that
\begin{equation*}
    W_2(\mu,\nu)^2 =\inf_{X_t}  E\bigg[ \int_0^1 \|f(X_t,t)\|_2^2 dt \bigg] + \lambda D(\rho_1,\nu) \, \text{ s.t. }\, X_0\sim\mu.
\end{equation*}
Moreover, if $X_t$ defined on the embedded space $\Z$, then $W_2$ is equivalent to the Wasserstein with geodesic distance $W_2(\mu,\nu)\simeq W_{d_\mathcal{M}^{2\alpha}}(\mu,\nu)$.
\end{customthm}
\begin{proof}
We recall that
\begin{equation*}
        W_2(\mu,\nu)^2 = \inf_{X_t}  \mathbb{E}\bigg[ \int_0^1 \|f(X_t,t)\|_2^2 dt \bigg] \, \text{ s.t. }dX_t = f(X_t,t)dt,\, X_0\sim\mu,\,X_1\sim\nu,
\end{equation*}
is equivalent to 
\begin{equation*}
    W_2(\mu, \nu)^2 = \inf_{(\rho_t,v)}\int_{0}^{1}\int_{\real^k} \|v(x,t)\|^2 \rho_t(dx) dt,
\end{equation*}
with the three constraints 
\begin{equation*}
    \textit{a)}\,\partial_t \rho_t + \nabla \cdot (\rho_t v) = 0 \quad \textit{b)}\, \rho_0 = \mu, \quad \textit{c)}\, \rho_1=\nu.
\end{equation*}
\citet{tong_trajectorynet_2020} (Thm. 4.1), showed that, for large $\lambda>0$ and $\rho_t$ satisfying \textit{a)}, this minimization problem is equivalent to
\begin{equation*}
        W_2(\mu, \nu)^2 = \inf_{(\rho_t,v)}\int_{0}^{1}\int_{\real^k} \|v(x,t)\|^2 \rho_t(dx) dt + \lambda KL(\rho_1\,||\nu),
\end{equation*}
where KL is the Kullback–Leibler divergence, we note that their proof is valid for any dissimilarity $D(\rho_1\,||\nu)$ respecting the identity of indiscernibles. Using the path formulation, by writing the integral as an expectation and taking the infimum over all absolutely continuous path, we have 
\begin{equation*}
    W_2(\mu,\nu)^2 =\inf_{X_t}  E\bigg[ \int_0^1 \|f(X_t,t)\|_2^2 dt \bigg] + \lambda D(\rho_1,\nu) \, \text{ s.t. }\,dX_t = f(X_t,t)dt,\, X_0\sim\mu.
\end{equation*}

Assuming the encoder $\phi$ achieves a loss of zero, then from Corr.~\ref{cor:zeroloss_geo} for $\alpha\in(0,1/2)$, we have $\|\phi(x_i)-\phi(x_j)\|_2 \simeq d^{2\alpha}_\M(x_i,x_j)$ for all $x_i,x_j\in\X\subseteq\M$, and sufficiently large $n$ and $K$. That is, there exist $c,C>0$ such that $c \,d_\M(x_i,x_j) \leq  \|\phi(x_i)-\phi(x_j)\|_2 \leq C d_\M(x_i,x_j)$ for all $x_i,x_j\in \X$. Then, for all $\pi\in\Pi(\mu,\nu)$, we have
\begin{equation*}
    c^p\int_{\X\times\X} d^{2\alpha}_\M(x,y)^p \pi(dx,dy)\leq \int_{\X\times\X} \|\phi(x)-\phi(y)\|_2^p\pi(dx,dy)\leq C^p\int_{\X\times\X}d^{2\alpha}_\M(x,y)^p\pi(dx,dy),
\end{equation*}
and taking the infimum with respect to $\pi\in\Pi(\mu,\nu)$ yields the desired result $W_2(\mu,\nu)\simeq W_{d_\mathcal{M}^{2\alpha}}(\mu,\nu)$.
\end{proof}


\subsection{Schr\"odinger bridge}
Let $\mathcal{D(\mu,\nu)}$ the space of probability distributions on $\mathcal{C}:=C([0,1],\real^k)$, with initial and final distribution $\mu$ and $\nu$, and $N$ the Wiener measure on $\mathcal{C}$, then the Schr\"odinger bridge (SB) problem is to find the time-evolving distribution $\rho_t$ such that
\begin{equation*}
    \min \mathrm{KL}(\rho_t\,||N) \text{ subject to } \rho_t \in\mathcal{D}(\mu,\nu),
\end{equation*}
where $\mathrm{KL}$ is the Kullback–Leibler divergence. It also admits a static formulation, i.e. minimizing with respect to $\pi\in\Pi(\mu,\nu)$. Using the measure $N_x^y$ over a Brownian bridge with initial and final condition $x$ and $y$, and $\rho_x^y$ defined similarly for $\rho\in\mathcal{D(\mu,\nu)}$, we can write $KL(\rho_t\,||N) = KL(\pi\,||\pi^N) + KL(\rho_x^y\,||N_x^y)$, where $\pi^N$ is the joint distribution between the initial and final states under the Wiener measure $N$. By choosing $\rho_x^y=N_x^y$, the \emph{static formulation} of the Schr\"odinger bridge is 
\begin{equation*}
   \min \mathrm{KL}(\pi\,||\pi^N) \text{ subject to } \pi \in\Pi(\mu,\nu).
\end{equation*}
When the Wiener measure has variance $\sigma^2$, the static Schr\"odinger bridge is equivalent to minimizing 
\begin{equation*}
    \inf_{\pi\in\Pi(\mu,\nu)} \int_{\real^k\times\real^k}(1/2)||x-y||_2^2\pi(dx,dy) + \sigma^2 H(\pi),
\end{equation*}
where $H$ is the entropy (see \cite{pavon2021data} and references therein). That is an optimal transport problem with entropic regularization, the same formulation as the Sinkhorn divergence \citeSupp{cuturi_sinkhorn_2013}, which can be solved with Sinkhorn's algorithm \citeSupp{sinkhorn1964relationship}. We compare our method with Diffusion Schrödinger Bridge (DSB)~\cite{de2021diffusion}, which solve the SB problem with an approximation of the iterative proportional fitting algorithm.

\section{Experiment details}\label{sec: exp_details}
For all experiments, we used the \textit{Runge-Kutta} RK4 ODE solver, for the density loss we used $h=0.01$, and $5$ nearest neighbors. The ODE network consists of three layers, and we concatenate two extra dimensions to the input as well as the time index. The encoder network is three layers with ReLU activation functions in between layers. For each epoch (local or global), we sampled $20$ batches of a given sample size without replacement. For both networks, we optimize with AdamW~\citeSupp{loshchilov2018decoupled} with default parameters. In practice, we did not use the energy loss, as our trajectories were already smooth, and it requires more careful parametrization (see Sec.~\ref{sec: additional_res}). For all experiments, we used TrajectoryNet with $1000$ iterations and \textit{whiten}. We used the authors' implementation of DSB with the \textit{basic} model and default parameters. 

\paragraph{Petal} For diffusion geodesic, we experiment with an RBF kernel with scale $0.1$ and $\alpha$-decay kernel (knn=5), both with maximum time scale $2^5$. The encoder consists of two linear layers of size $8$ and $32$, trained for $1000$ iterations. We used LeakyReLU in between layers of the ODE network, with hidden layers size $16,32,16$, and initial scales $\sigma_t=0.1$. We trained for $40$ local epoch with sample size $60$. We used $\lambda_d=35$ to weight the density loss and $\lambda_e=0.001$. For the hold-out experiments, we trained for $30$ local and $15$ global epoch with $\lambda_d=1$.

The diffusion parameters are generally small while training, for example when trained on the entire dataset $\sigma_t\in\{ 0.18,  0.15, -0.03,  0.11,  0.08\}$, and for the hold-out time three $\sigma_t\in\{0.08, 0.02, 0.16, 0.03, 0.08\}$.

\paragraph{Dyngen}
The diffusion geodesic was evaluated with an RBF kernel with scale $0.5$, and maximum time scale $2^5$. The encoder has two linear layers of size $8$ and $32$, trained for $1000$ iterations. We used CELU in between layers of the ODE network, with hidden layers size $16,32,16$, and initial scales $\sigma_t=0.2$. We trained for $50$ global epochs of size $60$, with $\lambda_d=5$. For the hold-out experiments, we used $\lambda_d=5$, and for $t=5$ we trained on $5$ local and $10$ global epochs. For $t=3$, since we have to hold-out the steps $(2,3)$ and $(3,4)$, we only trained using $50$ global epochs. 

Similar to the petal dataset, the diffusion parameters is empirically low, on the entire dataset it is $\sigma_t\in\{0.18, 0.20, 0.23, 0.18, 0.16\}$, and for the hold-out three $\sigma_t\in\{0.02, 0.08, 0.24, 0.22, 0.24\}$.

\paragraph{EB data}The EB data are publicly available\footnote{\url{https://data.mendeley.com/datasets/v6n743h5ng/1}}. For the genes trajectories we used the full geodesic autoencoder because we want to decode in PCA space, and then project back into gene space. The encoder consists of three layers of size $200,100,100$, and the decoder is $100,100,200$, both with ReLU activation between layers. We trained for $1000$ epochs with sample size of $100$ per time, with the $\alpha$-decay kernel ($5$ nearest neighbors, and max scale of $2^6$). This kernel is more expensive to compute since it relies on the approximation of the knn, but it circumvents the need to find a good scale as with the RBF kernel. The ODE network is three hidden layers of size $64$, with CELU activation in between layers. We trained for $20$ local and $10$ global epochs with sample size of $400$ per time. We used the density loss with weight $\lambda_d=20$.

\paragraph{AML} The AML dataset is publicly available\footnote{\url{https://www.ncbi.nlm.nih.gov/geo/query/acc.cgi?acc=GSE161676}}. We first embed the data in 50 dimensions PCA space. We use an encoder of size $50,8,8$ with ReLU activations in between layers, trained for $1000$ iterations with the RBF kernel (scale 0.5). For the trajectories we used an ODE network with three hidden layers of size $16,32,16$, with LeakyReLU in between layers, trained without noise for $50$ local epochs with $\lambda_d=20$.

\section{Additional Results}\label{sec: additional_res}

\begin{figure}[ht]
    \centering
    \includegraphics[width=1\linewidth]{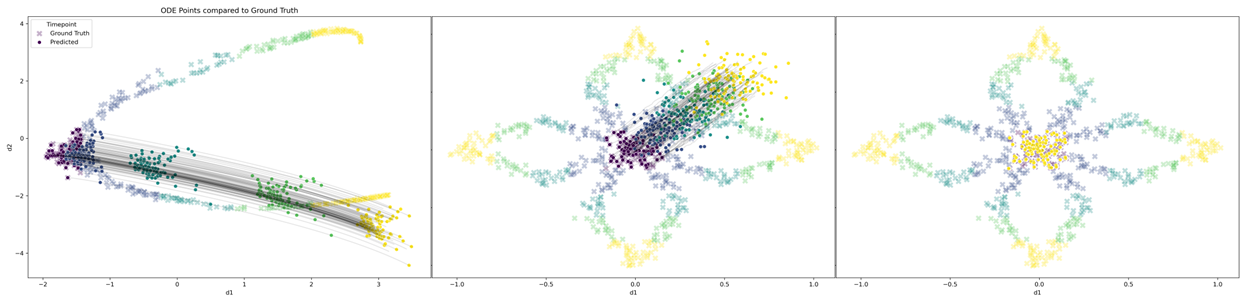}
    \caption{Trajectories with energy loss, and $\lambda_e=1$.}
    \label{fig:energy_loss}
\end{figure}

\begin{figure}[ht]
    \centering
    \includegraphics[width=0.8\linewidth]{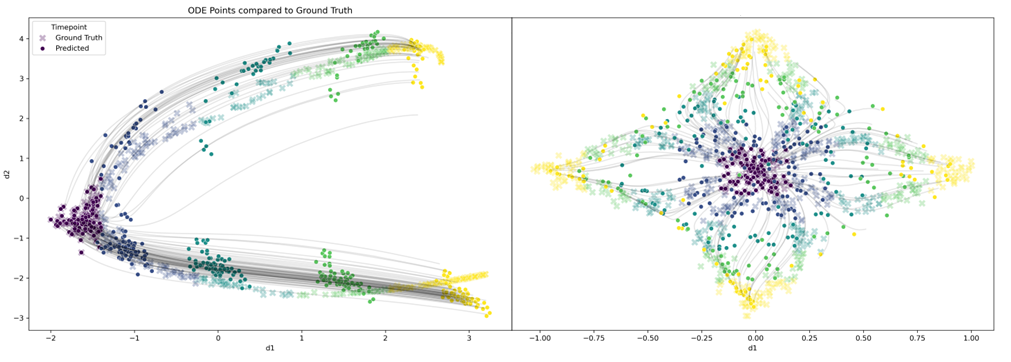}
    \caption{Trajectories with energy loss, and $\lambda_e=0.01$.}
    \label{fig:energy_loss_low}
\end{figure}

\paragraph{Energy loss} In Fig.~\ref{fig:energy_loss}, we present the trajectories using the same parametrization as detailed in Sec.~\ref{sec: exp_details} with the energy loss ($\lambda_e=1$). This loss requires much more memory since for each function evaluation we need to save the norm of the derivative. It should encourage smooth trajectory, but the accuracy of the trajectories is very sensitive to this loss. On Dyngen (left), it learns a straight line, and thus cannot bifurcate. On the petal, it initially learns a straight trajectory (middle, one epoch), and as training continues, it collapses in the center (right, ten epochs). Unless otherwise stated, we did not add it to our model since our trajectories were smooth. However, with careful parametrization, one could learn more accurate trajectories trajectories (see Fig.~\ref{fig:energy_loss_low}).

\paragraph{Ablation and hold-out}
To evaluate the prediction accuracy, we train by holding-out one timepoint, and we compute the distance between the held-out sample (ground truth) and the prediction. We use the $W_1$, the MMD with a Gaussian kernel, and the MMD with identity map ($L^2$ norm between the sample means). For the Petal and Dyngen datasets, we compute the Gaussian MMD by taking the average for the scales $0.1$ and $0.5$. For the EB data, we set the scale to $1500$.

\begin{figure}[ht]
    \centering
    \includegraphics[width=0.85\linewidth]{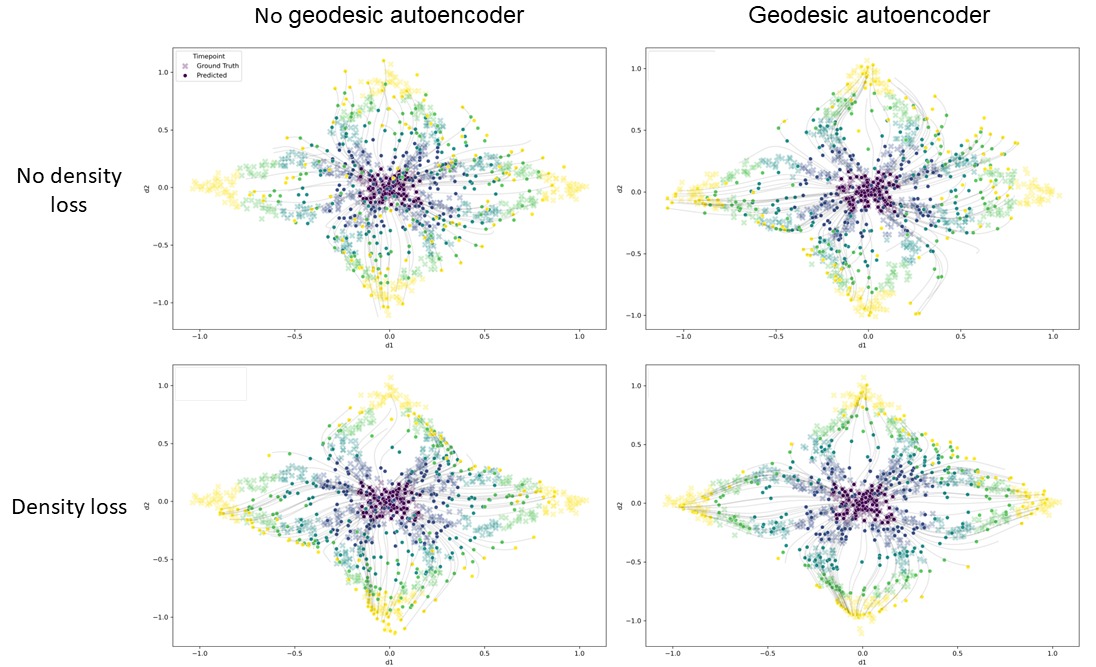}
    \caption{Petal dataset with and without the density loss (rows) or the geodesic autoencoder (columns).}
    \label{fig:petal_abla}
\end{figure}

\begin{figure}[ht]
    \centering
    \includegraphics[width=1\linewidth]{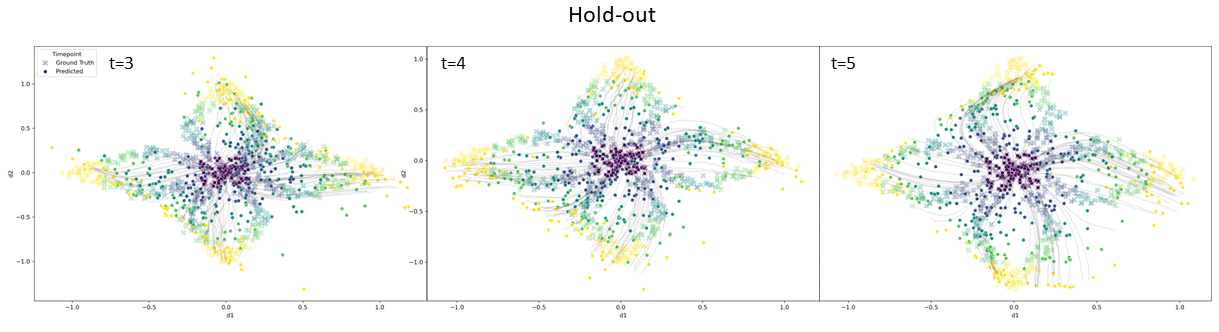}
    \caption{Trajectories for different hold-out time $t$ on the petal dataset.}
    \label{fig:petal_holdout}
\end{figure}

\begin{figure}[ht]
    \centering
    \includegraphics[width=0.85\linewidth]{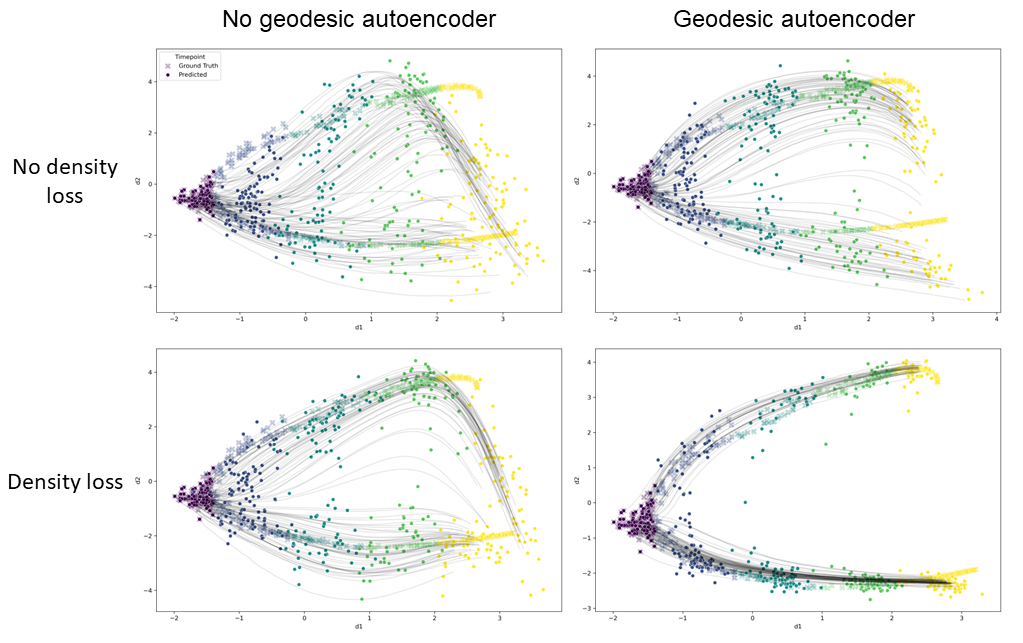}
    \caption{Dyngen dataset with and without the density loss (rows) or the geodesic autoencoder (columns).}
    \label{fig:dyngen_abla}
\end{figure}

\begin{figure}[ht]
    \centering
    \includegraphics[width=1\linewidth]{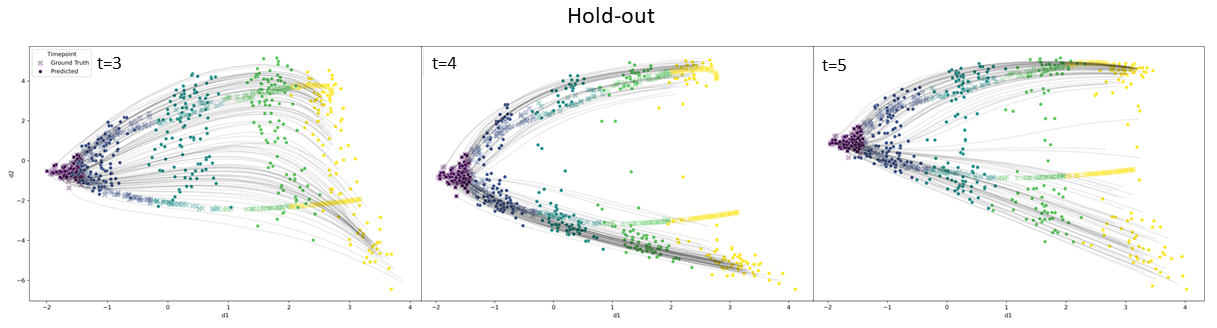}
    \caption{Trajectories for different hold-out time $t$ on the Dyngen dataset}
    \label{fig:dyngen_holdout}
\end{figure}

In Fig.~\ref{fig:petal_abla}, we present an ablation study for the density loss and the geodesic embedding. The parametrization was found for the model with density loss and embedding, both seem important to learn accurate trajectories. In Fig.~\ref{fig:petal_holdout}, we present the trajectories when we trained by holding one timepoint out both with the embedding and density loss. In Fig.~\ref{fig:dyngen_abla}, we reproduce the same ablation on Dyngen data. Here, the density loss is important for the trajectories to stay on the manifold, and the embedding appears to help with the bifurcation. To complement this figure, in Tab.~\ref{res: dyngen_traj_emb_vs_noemb}, we present the average of the KNN distance computed on the top $5\%$. On Dyngen, the trajectories with hold-out timepoints are presented in Fig.~\ref{fig:dyngen_holdout}. Holding out the timepoint $t=3$ seems to be the hardest to learn. In Tab.~\ref{tab:holdout_dyngen_petal}, we show the $W_1$ and MMD between the predicted values and ground truth, for MIOFlow, TrajectoryNet, and DSB. For the $W_1$, MIOFlow is always more accurate. In Tab~\ref{tab:eb_ho}, we compare the prediction accuracy on the EB data, depending on using the GAE or not. For the GAE, we consider the choice of two kernels when computing the distance $G$; the Gaussian kernel for various bandwidth parameters, and the $\alpha$-decay kernel for different $K$-nearest neighbors. Using the GAE improves the interpolation accuracy for all metrics. We embed the data in $200$ PCA, and train the autoencoder with layers $200,100,100,50$ for $1000$ epochs. We trained the Neural ODE for $80$ epochs with $\lambda_d = 20$, $\sigma_t = 0.2$, linear layers of size $16,32,16,50$ with CELU activation in between layers. We train by holding out timepoint $t$ for the GAE and the neural ODE. For the MMD with the Gaussian kernel, we set the scale to $1500$.

         

\begin{table}[htbp]
  \centering
  \caption{Leave-one-out $(t)$ $W_1$ and MMD with (G)aussian or (M)ean kernel between the predicted and ground truth point on the Petal and Dyngen datasets of MIOFlow, TrajectoryNet, and DSB. Lower is better.}
    \resizebox{\columnwidth}{!}{%
    \begin{tabular}{clccccccccc}
    \toprule
          &       & \multicolumn{3}{c}{$t=3$} & \multicolumn{3}{c}{$t=4$} & \multicolumn{3}{c}{$t=5$} \\
\cmidrule{3-11}          &       & $W_1$   & MMD(G) & MMD(M) & $W_1$   & MMD(G) & MMD(M) & $W_1$   & MMD(G) & MMD(M) \\
    \midrule
    \multirow{4}[2]{*}{Petal} & MIOFlow(ours) & \textbf{0.170} & \textbf{0.051} & 0.002 & \textbf{0.200} & \textbf{0.052} & 0.015 & \textbf{0.218} & \textbf{0.055} & 0.009 \\
          & TrajectoryNet & 0.379 & 0.568 & 0.001 & 0.347 & 0.341 & 0.003 & 0.264 & 0.096 & 0.004 \\
          & DSB   & 0.310 & 0.386 & \textbf{$<$0.001} & 0.260 & 0.087 & 0.029 & 0.441 & 0.159 & 0.081 \\
          & Baseline & 0.231 & 0.176 & 0.001 & 0.241 & 0.133 & \textbf{0.001} & 0.250 & 0.128 & \textbf{0.001} \\
    \midrule
    \multirow{4}[1]{*}{Dyngen} & MIOFlow(ours) & \textbf{0.509} & \textbf{0.181} & \textbf{0.022} & \textbf{1.787} & \textbf{0.324} & \textbf{3.071} & \textbf{1.450} & \textbf{0.491} & \textbf{1.719} \\
          & TrajectoryNet & 1.797 & 1.125 & 1.588 & 2.953 & 1.450 & 5.790 & 2.185 & 1.037 & 1.913 \\
          & DSB   & 0.767 & 0.260 & 0.070 & 2.699 & 0.583 & 6.521 & 2.823 & 1.083 & 5.546 \\
          & Baseline & 1.828 & 0.958 & 2.435 & 2.198 & 1.131 & 3.368 & 2.221 & 1.133 & 3.139 \\
          \bottomrule
    \end{tabular}%
    }
  \label{tab:holdout_dyngen_petal}
\end{table}%

\begin{table}[htbp]
  \centering
  \caption{Leave-one-out $(t)$ $W_1$ and MMD with (G)aussian and (M)ean kernel between the predicted and ground truth point on the EB datasets for MIOFlow with the Gaussian, $\alpha$-decay kernel, or without GAE. Lower is better.}
    \begin{tabular}{crcccccc}
    \toprule
          &       & \multicolumn{3}{c}{$t=2$} & \multicolumn{3}{c}{$t=3$} \\
\cmidrule{3-8}   GAE       &       & $W_1$   & MMD(G) & MMD(M) & $W_1$   & MMD(G) & MMD(M) \\
    \midrule
    \multirow{3}[1]{*}{$\alpha$-decay} & \multicolumn{1}{l}{knn $=10$} & 25.774 & 0.072 & 119.386 & \textbf{32.129} & 0.135 & 206.144 \\
          & \multicolumn{1}{l}{knn $=20$} & \textbf{25.697} & 0.064 & 104.579 & 32.729 & 0.143 & 228.890 \\
          & \multicolumn{1}{l}{knn $=30$} & 25.744 & 0.061 & \textbf{99.747} & 32.227 & 0.135 & 213.465 \\
          \midrule
    \multirow{3}[0]{*}{Gaussian} & \multicolumn{1}{l}{$\epsilon= 0.05$} & 26.873 & 0.064 & 115.905 & 32.844 & 0.132 & 213.900 \\
          & \multicolumn{1}{l}{$\epsilon=0.1$} & 26.107 & 0.064 & 109.653 & 32.776 & 0.123 & \textbf{203.617} \\
          & \multicolumn{1}{l}{$\epsilon=0.5$} & 26.318 & \textbf{0.059} & 101.969 & 32.963 & 0.138 & 223.679 \\
          \midrule
    No GAE & \multicolumn{1}{l}{} & 29.243 & 0.063 & 126.608 & 35.709 & 0.119 & 246.609 \\
    \midrule
    Baseline &       & 33.415 & 0.103 & 227.279 & 35.319 & \textbf{0.095} & 213.492 \\
    \bottomrule
    \end{tabular}%
  \label{tab:eb_ho}%
\end{table}%

\paragraph{Nearest neighbor distance}
To assist in evaluating how well the generated points land on the known manifold (ground truth points), we employ 1-NN distance from the generated points to the ground truth points. Table~\ref{tab:1nn} shows the 1-NN distance for several methods and datasets. Of note is the \textit{aggregate} column. Where aggregate is specified as \textit{mean}, we compute the mean 1-NN distance from the predicted point to the 1-NN at the same time index. Additionally, to help reduce sensitivity due to outliers, we computed these distances and report only the mean of the worst (highest 1-NN distance) quartile, where aggregate is specified as \textit{quartile}. 


\begin{table}[ht]
\caption{Average 1-NN distance of predicted points from a given method to ground truth points at the same time label across two datasets. Rows designated \textit{mean} are the average 1-NN distance, while those designated \textit{quartile} are the average 1-NN distance for the worst quartile. Lower is better.}
\centering
\bgroup
\def\arraystretch{1.5}%
\begin{tabular}{lcccl}
\toprule
                            & MIOFlow(ours) & TrajectoryNet   & DSB                          & aggregate  \\ \midrule
\multicolumn{1}{l|}{Petal} & \textbf{0.033} & 1.823 & \multicolumn{1}{l|}{0.286} & mean \\
\multicolumn{1}{l|}{Dyngen} & \textbf{0.586} & 2.562 & \multicolumn{1}{l|}{1.638} & mean \\ \midrule
\multicolumn{1}{l|}{Petal} & \textbf{0.096} & 2.769 & \multicolumn{1}{l|}{0.618} & quartile \\
\multicolumn{1}{l|}{Dyngen} & \textbf{1.163} & 4.277  & \multicolumn{1}{l|}{3.474} & quartile \\
\bottomrule
\end{tabular}
\egroup
\label{tab:1nn}
\end{table}

\begin{table}[htbp]
  \centering
  \caption{On the Dyngen dataset, average distance of the $10$-NN for predicted points between ground truth at the same time or any. The average is computed on the highest $5\%$ observations. Lower is better.}\label{res: dyngen_traj_emb_vs_noemb}
    \begin{tabular}{llcc}
    \toprule
    \textcolor[rgb]{ .114,  .11,  .114}{}GAE &       & time  & any \\
    \midrule
    \multicolumn{1}{c}{\multirow{3}[2]{*}{\textcolor[rgb]{ .114,  .11,  .114}{$\alpha$-decay}}} & knn$=5$ & \textcolor[rgb]{ .114,  .11,  .114}{19.922} & \textcolor[rgb]{ .114,  .11,  .114}{19.249} \\
          & knn$=10$ & \textcolor[rgb]{ .114,  .11,  .114}{19.987} & \textcolor[rgb]{ .114,  .11,  .114}{19.097} \\
          & knn$=15$ & \textcolor[rgb]{ .114,  .11,  .114}{18.579} & \textcolor[rgb]{ .114,  .11,  .114}{16.905} \\
    \midrule
    \multicolumn{1}{c}{\multirow{3}[2]{*}{\textcolor[rgb]{ .114,  .11,  .114}{Gaussian}}} & $\epsilon=0.05$ & \textcolor[rgb]{ .114,  .11,  .114}{16.245} & \textcolor[rgb]{ .114,  .11,  .114}{14.972} \\
          & $\epsilon=0.1$ & \textcolor[rgb]{ .114,  .11,  .114}{16.245} & \textcolor[rgb]{ .114,  .11,  .114}{15.491} \\
          & $\epsilon=0.5$ & \textcolor[rgb]{ .114,  .11,  .114}{\textbf{15.427}} & \textcolor[rgb]{ .114,  .11,  .114}{\textbf{14.101}} \\
    \midrule
    No GAE & Euclidean & \textcolor[rgb]{ .114,  .11,  .114}{21.589} & \textcolor[rgb]{ .114,  .11,  .114}{20.486} \\
    \bottomrule
    \end{tabular}%
  \label{tab:dyngen_manifold}%
\end{table}%

\begin{figure}[ht]
    \centering
    \includegraphics[width=0.85\linewidth]{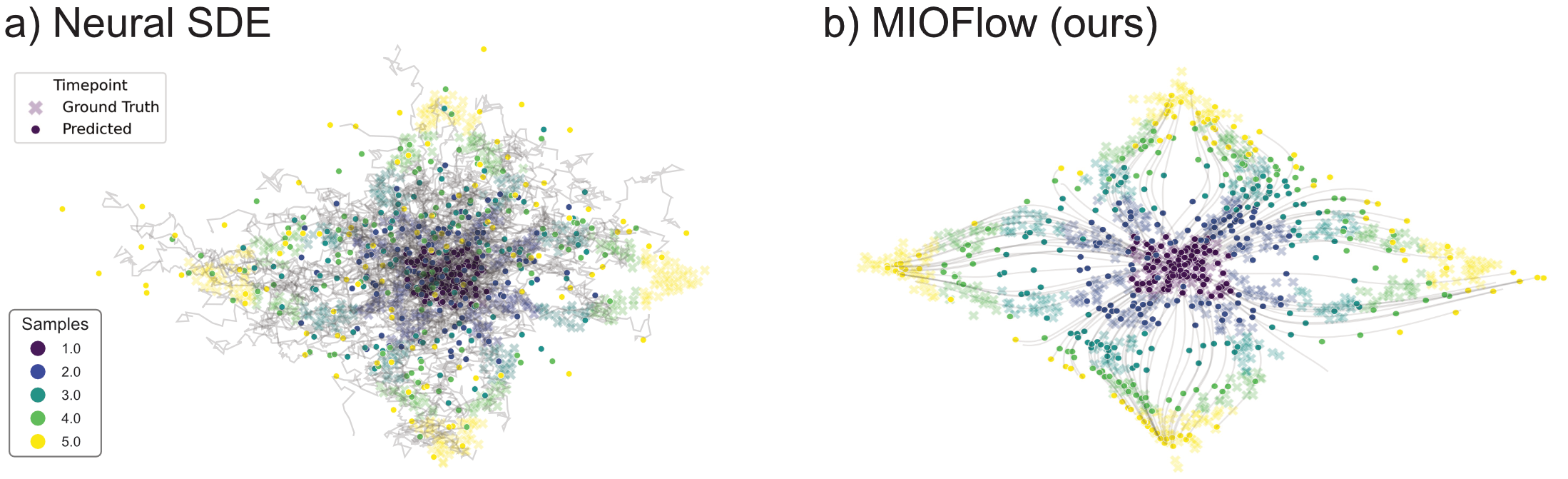}
    \caption{Comparison of Neural SDE (a) to MIOFlow (b) on the petal dataset.}
    \label{fig:neural-sde-petal}
\end{figure}
\begin{figure}[ht]
    \centering
    \includegraphics[width=0.85\linewidth]{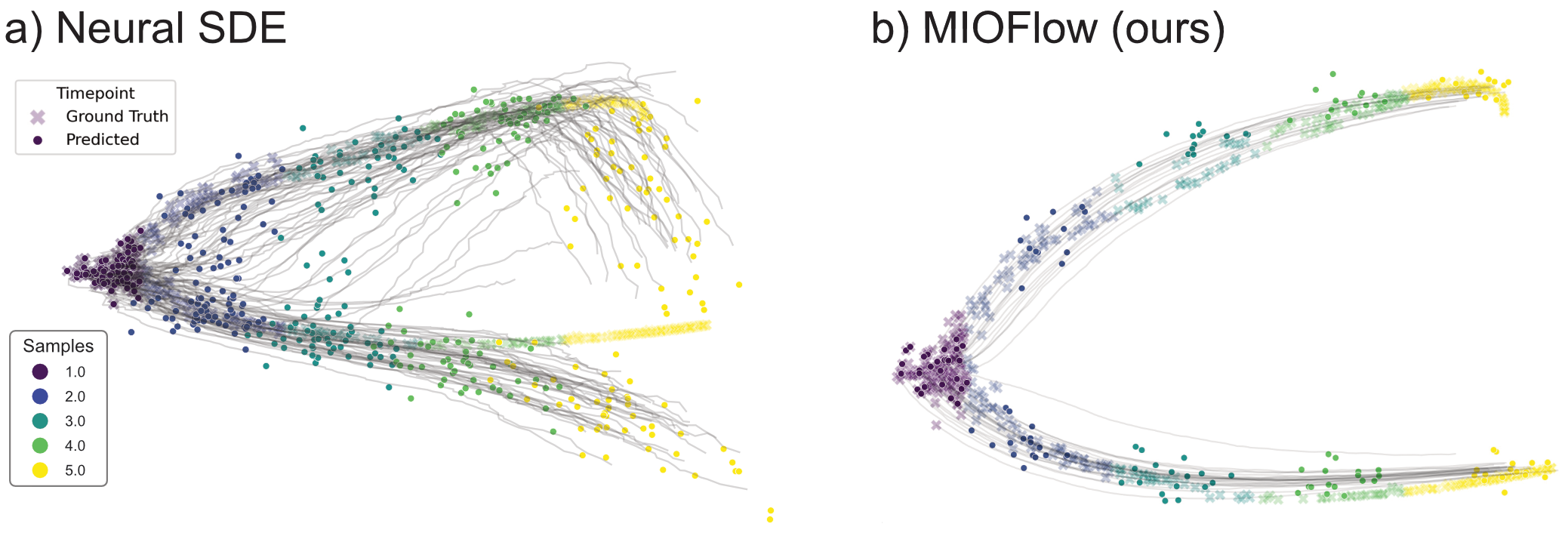}
    \caption{Comparison of Neural SDE (a) to MIOFlow (b) on the dyngen dataset.}
    \label{fig:neural-sde-dyngen}
\end{figure}

\paragraph{Neural SDE}
Additionally, we employed a Neural SDE from the PyTorch implementation of differentiable SDE Solvers~\citeSupp{li2020scalable,kidger2021neuralsde}, and compared its results to MIOFlow (Fig.~\ref{fig:neural-sde-petal} \& \ref{fig:neural-sde-dyngen}). We trained with a fixed diffusion function, but due to extensive training time (approximately two hours for 20 epochs for each dataset) we were unable to perform extensive hyperparameter optimization.



\begin{figure}[ht]
    \centering
    \includegraphics[width=0.85\linewidth]{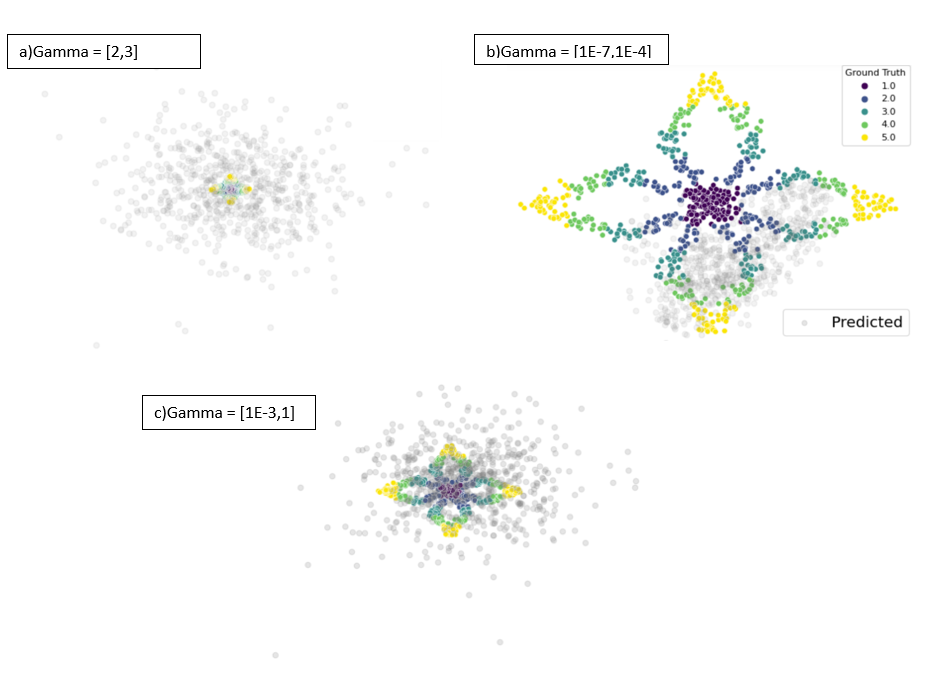}
    \caption{Comparison of DSB for different values of the noise scale $\gamma$.}
    \label{fig:DSB-gamma-petal}
\end{figure}

\paragraph{DSB on petal} In Fig.~\ref{fig:DSB-gamma-petal}, we show prediction of the DSB algorithm for different ranges of noise scale $\gamma$. With too little noise, the prediction appear to collapse on a subset of the data, while the larger scales are inaccurate for this dataset.

\section{Software Versions}
In Tab.~\ref{tab:gpus}, we present the GPU used during this work, and the following software version were used: \texttt{cudatoolkit==11.1.1, graphtools==1.5.2, numpy==1.19.5, phate==1.0.7 pytorch==1.10.2, pot==0.7.0, scikit-learn==0.24.2, scipy==1.5.3, scprep==1.1.0, torchdiffeq==0.2.2, torchsde==0.2.5}

\begin{table}[h]
\caption{GPU specifications of High Performance Cluster. Results were generated over a swath of GPU generations and models.}
\resizebox{\columnwidth}{!}{%
\bgroup
\def\arraystretch{1.5}%

\begin{tabular}{llllllll}
\toprule
Count & CPU Type    & CPUs/Node & Memory/Node (GiB) & GPU Type  & GPUs/Node & vRAM/GPU (GB) & Node Features                                                             \\ \midrule
1     & 6240        & 36        & 370               & a100      & 4         & 40            & cascadelake, avx2, avx512, 6240, doubleprecision, common, bigtmp, a100    \\
6     & 5222        & 8         & 181               & rtx5000   & 4         & 16            & cascadelake, avx2, avx512, 5222, doubleprecision, common, bigtmp, rtx5000 \\
4     & 5222        & 8         & 181               & rtx3090   & 4         & 24            & cascadelake, avx2, avx512, 5222, doubleprecision, common, bigtmp, rtx3090 \\
8     & E5-2637\_v4 & 8         & 119               & gtx1080ti & 4         & 11            & broadwell, avx2, E5-2637\_v4, singleprecision, common, gtx1080ti          \\
2     & E5-2660\_v3 & 20        & 119               & k80       & 4         & 12            & haswell, avx2, E5-2660\_v3, doubleprecision, common, k80                  \\ \bottomrule
\end{tabular}%
\egroup
}
\label{tab:gpus}
\end{table}

\clearpage
\bibliographystyleSupp{plainnat}
{\small
\bibliographySupp{supp}
}

\end{document}